\documentclass[10pt,twocolumn,letterpaper]{article}

\usepackage{cvpr}              % To produce the CAMERA-READY version

\usepackage[accsupp]{axessibility}  % Improves PDF readability for those with disabilities.

\usepackage{graphicx}
\usepackage{amsmath}
\usepackage{amssymb}
\usepackage{booktabs}

\usepackage[pagebackref=true,breaklinks=true,letterpaper=true,colorlinks,bookmarks=false, citecolor=ForestGreen]{hyperref}

\usepackage[capitalize]{cleveref}
\crefname{section}{Sec.}{Secs.}
\Crefname{section}{Section}{Sections}
\Crefname{table}{Table}{Tables}
\crefname{table}{Tab.}{Tabs.}
\crefformat{section}{\S#2#1#3} % see manual of cleveref, section 8.2.1
\crefformat{subsection}{\S#2#1#3}
\crefformat{subsubsection}{\S#2#1#3}
\crefformat{figure}{Figure~#2#1#3}

\usepackage{times}
\usepackage{epsfig}
\usepackage{hhline}
\usepackage[inline]{enumitem}
\usepackage[normalem]{ulem}
\usepackage[dvipsnames]{xcolor}

\usepackage[square,numbers,sort&compress]{natbib}
\usepackage{algpseudocode}
\usepackage[font=small,labelfont=bf]{caption}
\usepackage{array}
\usepackage{multirow}
\usepackage{algorithm}
\usepackage{subcaption}
\usepackage{xparse}
\usepackage{pifont}
\usepackage{bm}

\usepackage{listings}
\usepackage{mwe} % for placeholder images
\usepackage{makecell}
\usepackage{color, colortbl}
\usepackage[normalem]{ulem} % strikethrough font

\makeatletter
\def\ps@myheadings{%
    \let\@oddfoot\@empty\let\@evenfoot\@empty
    \def\@evenhead{\thepage\hfil\slshape\leftmark}%
    \def\@oddhead{{\slshape\rightmark}\hfil\thepage}%
    \let\@mkboth\@gobbletwo
    \let\sectionmark\@gobble
    \let\subsectionmark\@gobble
    }
  \if@titlepage
  \renewcommand\maketitle{\begin{titlepage}%
  \let\footnotesize\small
  \let\footnoterule\relax
  \let \footnote \thanks
  \null\vfil
  \vskip 60\p@
  \begin{center}%
    {\LARGE \@title \par}%
    \vskip 3em%
    {\large
     \lineskip .75em%
      \begin{tabular}[t]{c}%
        \@author
      \end{tabular}\par}%
      \vskip 1.5em%
    {\large \@date \par}%       % Set date in \large size.
  \end{center}\par
  \@thanks
  \vfil\null
  \end{titlepage}%
  \setcounter{footnote}{0}%
}
\else
\renewcommand\maketitle{\par
  \begingroup
    \renewcommand\thefootnote{\@fnsymbol\c@footnote}%
    \def\@makefnmark{\rlap{\@textsuperscript{\normalfont\@thefnmark}}}%
    \long\def\@makefntext##1{\parindent 1em\noindent
            \hb@xt@1.8em{%
                \hss\@textsuperscript{\normalfont\@thefnmark}}##1}%
    \if@twocolumn
      \ifnum \col@number=\@ne
        \@maketitle
      \else
        \twocolumn[\@maketitle]%
      \fi
    \else
      \newpage
      \global\@topnum\z@   % Prevents figures from going at top of page.
      \@maketitle
    \fi
    \thispagestyle{plain}\@thanks
  \endgroup
  \setcounter{footnote}{0}%
}
\makeatother
\newcommand{\hl}[1]{{\textbf{#1}}}
\definecolor{Gray}{gray}{0.9}
\newcolumntype{g}{>{\columncolor{Gray}}c}
\newcolumntype{?}{!{\vrule width 1pt}}

\def\fig#1{Fig.~\ref{fig:#1}}
\def\imw#1#2{\includegraphics[width=#2\linewidth]{#1.png}}
\def\imwjpg#1#2{\includegraphics[width=#2\linewidth]{#1.jpg}}

\def\imwh#1#2#3{\includegraphics[width=#2\linewidth,height=#3\textheight]{#1.png}}

\newcommand{\tb}[3]{\setlength{\tabcolsep}{#2mm}\begin{tabular}{#1}#3\end{tabular}}
\newcommand{\ol}[3]{\begin{#1}[leftmargin=*,topsep=3pt]\setlength{\itemsep}{#2mm}#3\end{#1}}

\def\lossF{\mathcal{L}_f}
\def\lossG{\mathcal{L}_g}

\newcommand{\algorithmName}{Hierarchical Segment Grouping\xspace}
\newcommand{\algorithmShort}{HSG\xspace}

\graphicspath{
{figs/},
{figs/semantic/},
{figs/hierarchy/},
{figs/context/},
{figs/attention/},
{figs/cosegmentation/},
}

\def\cvprPaperID{353} 
\def\confName{CVPR}
\def\confYear{2022}

\begin{document}

\title{
Unsupervised Hierarchical Semantic Segmentation with\\ Multiview Cosegmentation and Clustering Transformers
}

\author{
\tb{@{}ccccc@{}}{4}{
Tsung-Wei Ke&
Jyh-Jing Hwang& 
Yunhui Guo& 
Xudong Wang& 
Stella X. Yu
}\\
%{\tt\small \{twke,yunhui,xdwang,stellayu\}@berkeley.edu, jyh@gmail.com}\\
%{\tt\small \{twke,jyh,yunhui,xdwang,stellayu\}@berkeley.edu}\\
UC Berkeley / ICSI
}
\maketitle

\begin{abstract}

Unsupervised semantic segmentation aims to discover groupings within and across images that capture object- and view-invariance of a category without external supervision.  Grouping naturally has levels of granularity, creating ambiguity in unsupervised segmentation.  Existing methods avoid this ambiguity and treat it as a factor outside modeling, whereas we embrace it and desire hierarchical grouping consistency for unsupervised segmentation.

We approach unsupervised segmentation as a pixel-wise feature learning problem.  Our idea is that a good representation shall reveal not just a particular level of grouping, but any level of grouping in a consistent and predictable manner.  We enforce spatial consistency of grouping and bootstrap feature learning with co-segmentation among multiple views of the same image, and enforce semantic consistency across the grouping hierarchy with clustering transformers between coarse- and fine-grained features.

We deliver the first data-driven unsupervised hierarchical semantic segmentation method called \algorithmName (\algorithmShort). Capturing visual similarity and statistical co-occurrences, \algorithmShort also outperforms existing unsupervised segmentation methods by a large margin on five major object- and scene-centric benchmarks. 
%Cityscapes ($+6.1\%$), PASCAL VOC ($+6.8\%$), COCO-Stuff ($+7.4\%$), Potsdam ($+8.8\%$), and KITTI-STEP ($+2.5\%$).

\end{abstract}

\def\rowImg#1{
\imw{img_#1}{0.245}&
\imw{revisit_#1}{0.245}&
\imw{segsort_#1}{0.245}&
\imw{hsg_#1}{0.245}\\[-2pt]
}
\def\figTeaser#1{
    \begin{figure}[#1]
        \centering
        \tb{c}{0}{
        % Hierarchical segmentation
        \imw{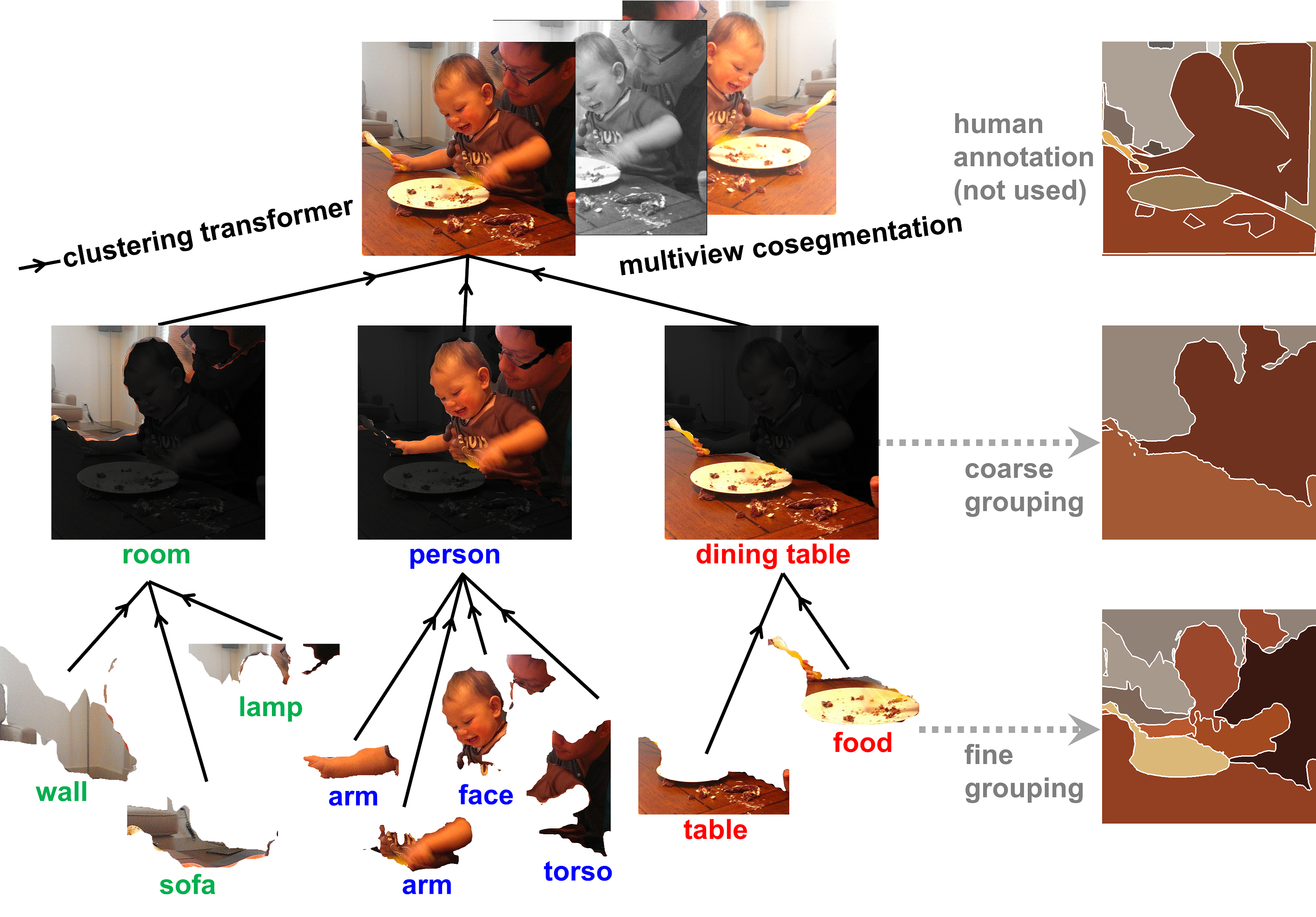}{1}\\
        %\imw{teaserFigure}{1}\\
        \midrule
        % Semantic segmentation
        \tb{@{}cccc@{}}{0.2}{%
        \rowImg{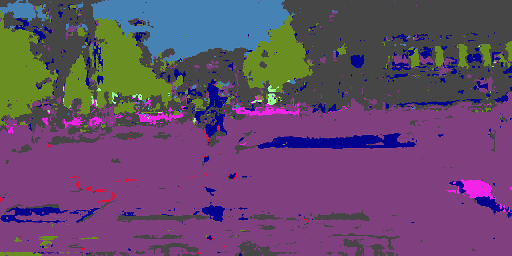}
        \rowImg{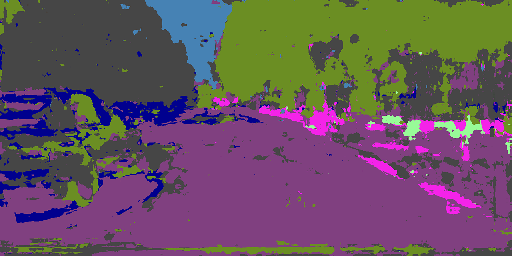}
        image & 
        Revisit~\cite{van2021revisiting}& 
        SegSort~\cite{hwang2019segsort}& 
        Our \algorithmShort\\
        }
        }
        \caption{
        We develop an unsupervised semantic segmentation method by embracing the ambiguity of grouping granularity and desiring hierarchical grouping consistency for unsupervised segmentation.  {\bf Top:} We formulate it as a pixel-wise feature learning problem, such that a good feature must be able to best reveal any level of grouping in a consistent and predictable manner.  We bootstrap feature learning from multiview cosegmentation and enforce grouping consistency with clustering transformers.  {\bf Bottom: } 
        Our method can not only deliver {\it hierarchical} semantic segmentation, but also outperform the state-of-the-art unsupervised segmentation methods by a large margin.  Shown are sample Cityscapes results.}
        \label{fig:teaser}
    \end{figure}
}

\section{Introduction}
\label{sec:intro}

\figTeaser{t}

Semantic segmentation requires figuring out the semantic category for each pixel in an image.  Learning such a segmenter from unlabeled data is particularly challenging, as neither pixel groupings nor semantic categories are known.

If pixel groupings are known, semantic segmentation is reduced to an unsupervised image (segment) recognition problem, to which contrast learning methods  \cite{wu2018unsupervised,he2020momentum,chen2020simple,wang2021unsupervised} could apply, on computed segments instead of images.

If semantic categories are known, semantic segmentation is reduced to a weakly supervised segmentation problem with coarse annotations of image-level tags;  pixel labeling can be predicted from image classifiers \cite{kolesnikov2016seed,ke2021spml}.

The fundamental task of unsupervised semantic segmentation is {\it grouping}, not {\it semantics} in terms of {\it naming}, which is unimportant other than the convenience of tagging segments in the same or different groups.
The challenge of unsupervised semantic segmentation is to discover groupings within and across images that capture object- and view-invariance of a category without external supervision, so that (Fig.~\ref{fig:teaser}):
{\bf 1)} A baby's face and body are parts of a whole in the same image; %
{\bf 2)} The whole baby is separated from the rest of the image; %
%{\bf 3)} The same baby appearing in different images with various poses, viewing conditions, and environments can be identified as the same instance;
{\bf 3)} A baby instance is more similar to another baby instance than to a cat instance, despite their different poses, illuminations, and backgrounds.

Several representative approaches have been proposed for tackling this challenge under different assumptions. 
\ol{itemize}{-0.5}{
\item{\bf Visual similarity:} SegSort~\cite{hwang2019segsort} first partitions each image into segments based on contour cues and then by  segment-wise contrastive learning discovers clusters of visually similar segments.  However, semantics by visual similarity is far too restrictive: A semantic whole is often made up of visually dissimilar parts.  Parts of {\it body} such as {\it head} and {\it torso} look very different; it is not their visual similarity but their spatial adjacency and statistical co-occurrence that bind them together.
\item {\bf Spatial stability:} IIC~\cite{ji2019invariant} maximizes the mutual information between clusterings from two views of the same image related by a known spatial transformation, enforcing stable clustering while assuming that a fixed number of clusters are equally likely within an image.  It works best for coarse and balanced texture segmentation and has major trouble scaling up with the scene complexity.
\item{\bf Image-wise feature learning:} 
~\cite{wang2021dense,van2021revisiting} train representations on object-centric datasets with multiscale cropping to sharpen the representation within the image.  These methods do not work well on scene-centric datasets where an image has more than one dominant semantic class.
}

Grouping as well as semantics naturally have different levels of granularity: A {\it hand} is an articulated configuration of a {\it palm} and five {\it fingers}, likewise a {\it person} of a {\it head}, a {\it torso},  two {\it arms}, and two {\it legs}.  Such an inherent grouping hierarchy poses a major challenge: Which level should an unsupervised segmentation method target at and what is the basis for such a determination?   Existing methods avoid this ambiguity and treat it as either a factor outside the segmentation modeling, or an aspect of secondary concern.

Our key insight is that the inherent hierarchical organization of visual scenes is not a nuisance for scene parsing, but a universal property that we can exploit and desire for unsupervised segmentation. This idea has previously led to a general image segmenter that handles texture and illusory contours through edges entirely without any explicit characterization of texture or curvilinearity \cite{yu:scale05}.  We now advance the concept to data-driven representation learning: A good representation shall reveal not just a particular level of grouping, but any level of grouping in a consistent and predictable manner across different levels of granularity.

We approach unsupervised semantic segmentation as an unsupervised pixel-wise feature learning problem.  Our objective is to best produce a consistent hierarchical segmentation for each image in the entire dataset based entirely on hierarchical clusterings in the feature space  (Fig.~\ref{fig:teaser}).  
Specifically, given the pixel-wise feature, we perform hierarchical groupings {\it within} and {\it across} images and their transformed versions (i.e.,{\it views}).   In turn, groupings at each level impose a desire on how the feature should be improved to maximize the discrimination among different groups.

%Specifically, we start from first creating the bottom level of the hierarchy, with visual elements that respect the low-level contour cues as in SegSort~\cite{hwang2019segsort}.
%
%We then build up higher levels of the hierarchy by formulating a clustering transformer~\cite{vaswani2017attention,tsitsulin2020graph}, which merges segments by consulting their feature similarities and global context.
%
%We transform the images into multiple views and train the network to predict consistent cosegmentations at every level.

Our model has two novel technical components: {\bf 1) Multiview cosegmentation} is to not only enforce spatial consistency between segmentations across views, but also bootstrap feature learning from visual similarity and co-occurrences in a simpler clean setting;  {\bf 2) Clustering transformers} are used to enforce semantic consistency across different levels of the feature grouping hierarchy. 

To summarize, our work makes three contributions.
\ol{enumerate}{-0.5}{
\item {\bf We deliver  the first unsupervised hierarchical semantic segmentation} method that can produce parts and wholes in a data-driven manner from an arbitrary collection of images, whether they come from object-centric or scene-centric datasets.

\item {\bf We are the first to embrace the ambiguity of grouping granularity} and exploit the inherent grouping hierarchy of visual scenes to learn a pixel-wise feature representation for unsupervised segmentation.  It can thus discover semantics based on not only visual similarity but also statistical co-occurrences.

\item {\bf We outperform existing unsupervised (hierarchical) semantic segmentation methods by a large margin} on not only object-centric but also scene-centric datasets.
}

\section{Related Work}
\label{sec:work}

\noindent \textbf{Image segmentation}
refers to the task of partitioning an image into visually coherent regions.  Traditional approaches often consist of two steps: extracting local features and clustering them based on different criteria, \eg,  mode-finding \cite{comaniciu2002mean,banerjee2005clustering}, or graph partitioning \cite{felzenszwalb2004efficient,shi2000normalized,malik2001contour,stella2003multiclass,yu2004segmentation}. 

\noindent
{\bf Hierarchical image segmentation} has been supervisedly learned from how humans perceive the organization of an image ~\cite{arbelaez2010contour}: While each individual segmentation targets a  particular level of grouping, the collection of individual segmentations present the perceptual hierarchy statistically.  

A typical choice for representing a hierarchical segmentation is  contours: They are first detected to sharply localize region boundaries ~\cite{hwang2015pixel,xie2015holistically} and can then be removed one by one to reveal coarser segmentations (OWT-UCM ~\cite{arbelaez2010contour}).  

Such models are trained on individual ground-truth segmentations, hoping that coarse and fine-grained organization would emerge automatically from common and rare contour occurrences respectively in the training data.

In contrast, our model is trained on 
multi-level segmentations unsupervisedly discovered by feature clustering, and it also operates directly on segments instead of contours.

\noindent
{\bf Semantic segmentation} refers to the task of partitioning an image into regions of different semantic classes.  Most deep learning models treat segmentation as a spatial extension of image recognition and formulate it as a pixel-wise classification problem.  They are often based on Fully Convolutional Networks \cite{lecun1989backpropagation,long2015fully,chen2016deeplab}, incorporating information from multiple scales \cite{chen2017rethinking, he2004multiscale,shotton2009textonboost,kohli2009robust,ladicky2009associative,gould2009decomposing, yao2012describing, mostajabi2014feedforward,aaf2018,hwang2019adversarial,ke2021spml,hwang2021contextual}.  

SegSort~\cite{hwang2019segsort} does not formulate segmentation as pixel-wise labeling, but pixel-segment contrastive learning that operates directly on segments delineated by contours.  It learns pixel-wise features in a non-parametric way, {\it with} or {\it without} segmentation supervision.  SPML~\cite{ke2021spml} extends it to unify segmentation with various forms of weak supervision: image-level tags, bounding boxes, scribbles, or points.

\noindent 
\textbf{Unsupervised semantic segmentation} has been modeled by non-parametric methods using statistical features and graphical models ~\cite{russell2009segmenting,tighe2010superparsing,liu2011nonparametric}.  For example, ~\cite{russell2009segmenting} proposes to discover region boundaries by mining the statistical differences of matched patches in coarsely aligned images.  

There are roughly three lines of recent unsupervised semantic segmentation methods.
{\bf 1)} One way is to increase the location sensitivity of the feature learned from images ~\cite{wu2018unsupervised,he2020momentum,chen2020simple,wang2021unsupervised}, by either adding an additional contrastive loss between pixels based on feature correspondences across views ~\cite{wang2021dense}, or using stronger  augmentation and constrained cropping ~\cite{van2021revisiting,selvaraju2021casting}.
{\bf 2)} A pixel-level {\it feature} encoder can be learned directly by maximizing discrimination between pixels based on either contour-induced segments ~\cite{hwang2019segsort} or
 region hierarchies ~\cite{zhang2020self} derived from OWT-UCM ~\cite{arbelaez2010contour}. 
Segmentation is indicated by pixel feature similarity and semantic labels can be inferred from retrieved nearest neighbours in a labeled set.
{\bf 3)} A pixel-wise {\it cluster} predictor can be directly learned by maximizing the mutual information between cluster predictions on augmented views of the same instance at corresponding pixels ~\cite{ji2019invariant,ouali2020Auto}.
  
Our model advances pixel-wise feature learning methods ~\cite{hwang2019segsort,ke2021spml,zhang2018exfuse}: It contrasts features based on feature-induced hierarchical groupings themselves, and most strikingly, directly outputs consistent hierarchical segmentations.

\def\figFramework#1{
    \begin{figure*}[#1]
        \centering
        \vspace{-10pt}
        \includegraphics[width=\textwidth]{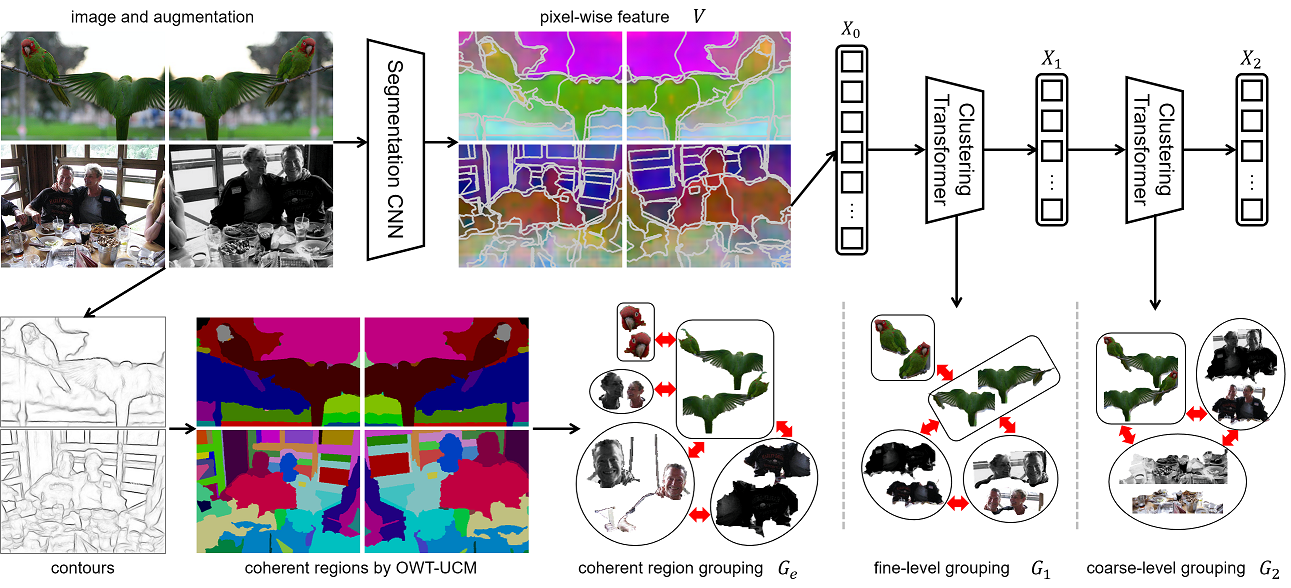}
        \caption{Our model consists of two essential components: {\bf 1)} multiview cosegmentation and {\bf 2)} hierarchical grouping.  We first produces pixel-wise feature $V$, from which we cluster to get base cluster feature $X_0$ and grouping $G_0$.  Each $G_0$ region is split w.r.t coherent regions derived by OWT-UCM procedure, which is marked by the white lines.  We create three groupings--$G_e$, $G_1$ and $G_2$ in multiview cosegmentation fashion.  We obtain $G_e$ by inferring the coherent region segmentation according to how each view is spatially transformed from the original image.  Starting with input $X_0$ of an image and its augmented views, we conduct feature clustering to merge $G_0$ into $G_1$, and then, $G_1$ into $G_2$.  Based on $G_e$, $G_1$ and $G_2$, we formulate a pixel-to-segment contrastive loss for each grouping.  Our \algorithmShort learns to generate discriminative representations and consistent hierarchical segmentations for the input images.}
        \label{fig:framework}
    \end{figure*}
}

\def\figFeatAndGroup#1{
    \begin{figure}[#1]
        \centering
        \vspace{-10pt}
        \includegraphics[width=\linewidth]{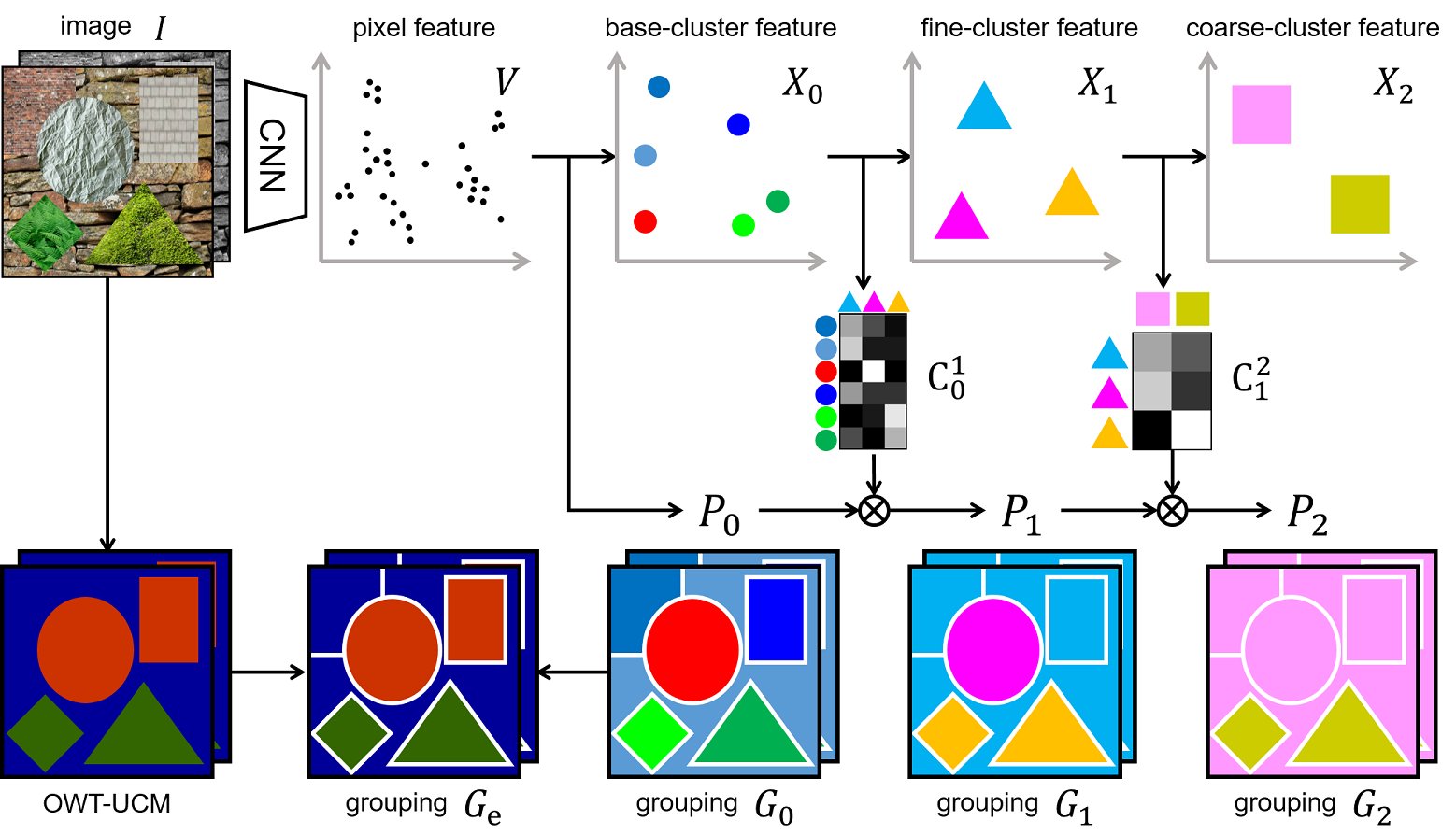}
        \caption{{\bf Method overview}.  We aim to learn a CNN that maps each pixel to a point in the feature space $V$ such that successively derived cluster features $X_0,X_1,X_2$ produce good and consistent hierarchical pixel groupings $G_e,G_1,G_2$.  
        Their consistency is enforced through clustering transformers $C_l^{l\!+\!1}$, which dictates how feature clusters at level $l$ map to feature clusters at level $l\!+\!1$.  Note that $G_0$ results from clusters of $V$, and $G_e$ from OWT-UCM edges.  $P_l$ is the probabilistic version of $G_l$, and $G_l$ the winner-take-all binary version of $P_l$; $P_0 \sim G_0$.
        For $l\!\ge\!0$, $P_{l+1}$ results from propagating
        $P_l$ by $C_l^{l\!+\!1}$.  Groupings $G_e,G_1,G_2$ in turn impose desired feature similarity and drive feature learning.  We co-segment multiple views of the same image to capture spatial consistency, visual similarity, statistical co-occurences, and semantic hierarchies.
        }
        \label{fig:featAndGroup}
    \end{figure}
}

\def\rowcs#1#2{
&
\imw{#1_edge#2}{0.249}&
\imw{#1_fine_tsne#2}{0.249}&
\imw{#1_coarse_tsne#2}{0.249}
\\[-2pt]
}
\def\rowCS#1#2{
\imw{#1_image#2}{0.249}&
\imw{#1_instance#2}{0.249}&
\imw{#1_finehrchy#2}{0.249}&
\imw{#1_coarsehrchy#2}{0.249}
\\[-2pt]
}
\def\figCoseg#1{
    \begin{figure}[#1]
        \centering
        \vspace{-10pt}
        \tb{@{}cccc@{}}{0.1}{
        \rowcs{2007_004481}{}
        image& 
        OWT-UCM %$\searrow$ 
        & 
        \multicolumn{2}{c}{
        \phantom{xxx}feature
        $\downarrow$\phantom{xxx}$\downarrow$ clustering}\\
        \rowCS{2007_004481}{}
        \rowCS{2007_004481}{_aug2}
        \rowCS{2007_004481}{_aug3}
        \multicolumn{2}{r}{coherent region $G_e$} & 
        fine $G_1$ &
        coarse $G_2$ \\
        }
        \caption{We co-segment multiple views (Column 1) of the same image by
        OWT-UCM edges ($G_e$, Column 2) or by feature clustering at fine and coarse levels ($G_1,G_2$, Columns 3-4).  
        White lines mark the segments derived from pixel feature clustering and OWT-UCM edges.  The color of feature points (pixels) mark grouping in the feature space (segmentation in the image) consistently across rows in the same column, per spatial transformations between views. 
        $G_2$'s coarse segmentations simply merge $G_1$'s fine segmentations, their consistency enforced by our clustering transformers.   Minimizing $\lossF(G_e),\lossF(G_1),\lossF(G_2)$ ensures respectively that our learned feature is grounded in low-level coherence, yet with view invariance, and capable of capturing semantics at multiple levels and producing hierarchical segmentations.
       }
        \label{fig:coseg}
    \end{figure}
}

\def\rowTS#1#2{
\imw{#1_prototypes#2}{0.333}&
\imw{#1_finehrchy_centroid#2}{0.333}&
\imw{#1_coarsehrchy_centroid#2}{0.333}
\\[-2pt]
}
\def\figHierTransformer#1{
    \begin{figure}[#1]
        \centering
        \vspace{-10pt}
        \tb{@{}ccc@{}}{0.1}{
        $X_0$ & $X_1$ & $X_2$\\
        \rowTS{2007_004481}{}
        }
        \tb{c}{0}{%
        \imwh{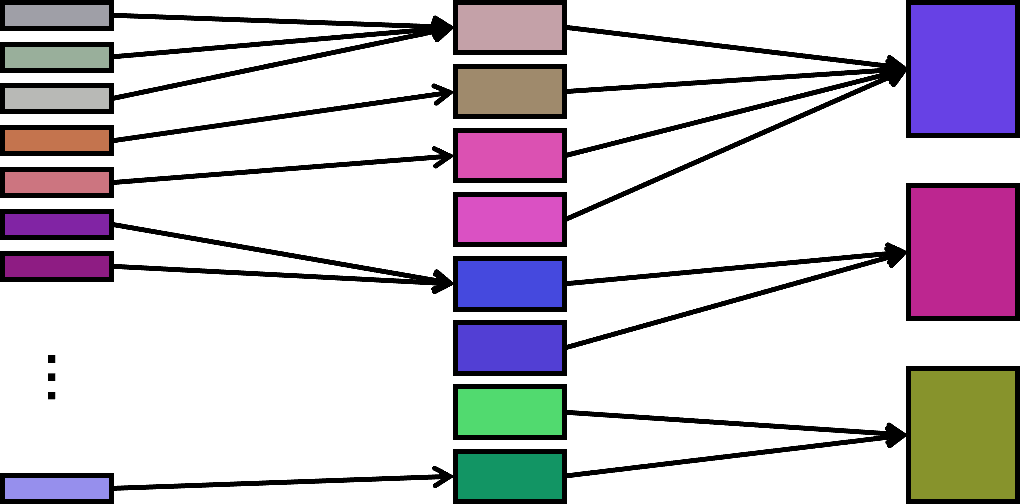}{0.75}{0.1}
        }
        \caption{Our clustering transformer hierarchically groups image segments in the learned feature space.  {\bf Top:} Base-cluster feature ($X_0$), fine- and coarse-level cluster features ($X_1$ and $X_2$) produced by our framework.  These high-dimensional features are jointly reduced to 3-dimension with tSNE.  White lines mark oversegmentation.  We remap these region-level features back to pixel-level by matching pixels with their segment and feature clustering index.  {\bf Bottom:} Low-to-high level cluster mappings predicted by our clustering transformer, whose colorings are consistent with top figures.  Each color represents a cluster (or segement) at corresponding level in the hierarchy.  We can infer hierarchical groupings of the image using our clustering transformer.}
        \label{fig:hierTransformer}
    \end{figure}
}

\def\figClusterTransformer#1{
    \begin{figure}[#1]
        \centering
        \includegraphics[width=\linewidth]{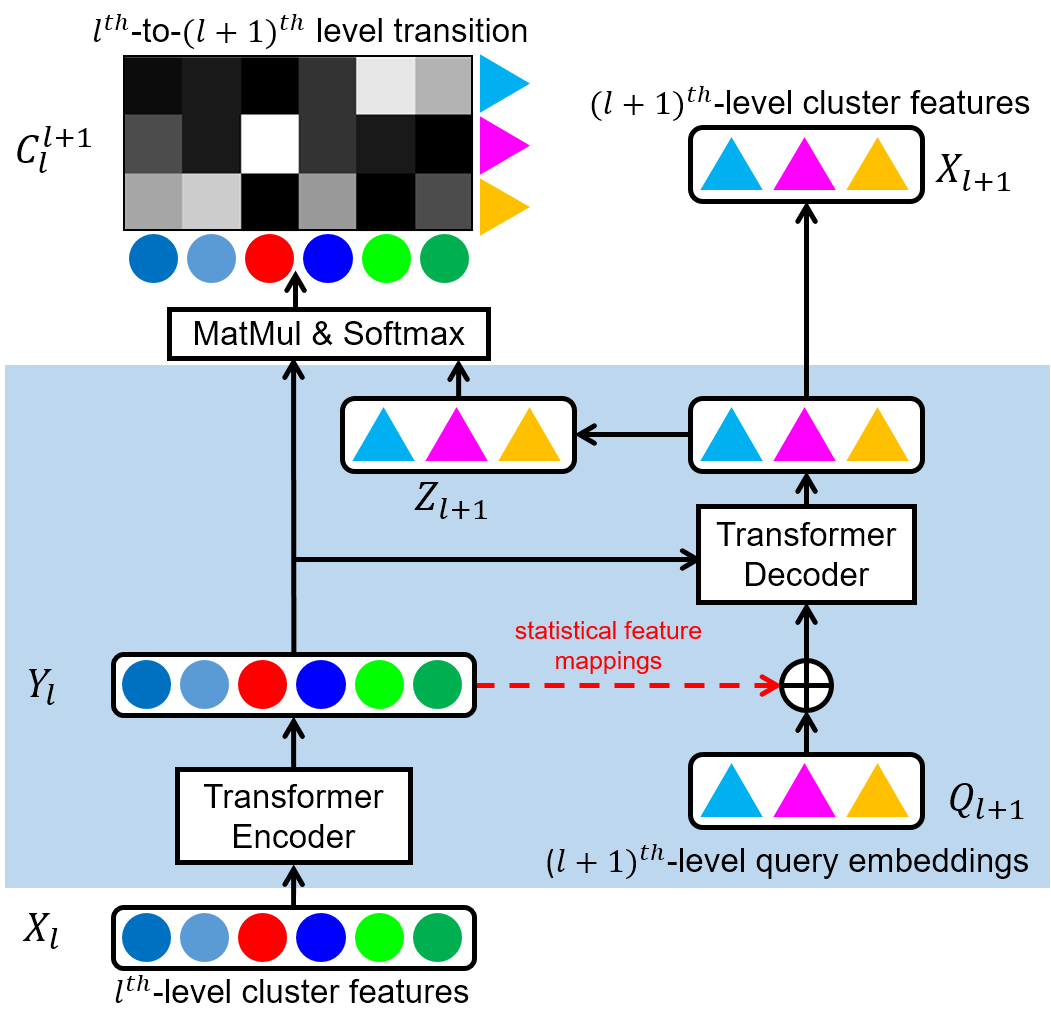}
        \caption{Our clustering transformer
        enforces grouping consistency across levels by mapping feature $X_l$ to $X_{l+1}$ with feature transition $C_l^{l+1}$.  $X_{l+1}$ and $C_l^{l+1}$ are learned simultaneously.
       Shown here for level $l\!=\!0$ in \fig{featAndGroup},
        the transformer encoder takes inputs $X_{l}$
         and outputs contextualized feature $Y_l$.
         The transformer decoder takes learnable inputs from query embeddings $Q_{l+1}$, and outputs $X_{l+1}$ and additionally projected feature $Z_{l+1}$.
         The transition is predicted as: $C_l^{l+1}\!=\!\text{softmax}\left(\frac{1}{\sqrt{m}}Y_l^\top Z_{l+1}\right)$; $m$ is the feature dimension.  {\color{red} Statistical feature mapping}:
         Calculate $Y_l$'s mean and std,  transform them by fc layers, and add to $Q_{l+1}$ for instance adaptation. }
        \label{fig:clusterTransformer}
    \end{figure}
}

\section{Hierarchical Segment Grouping (HSG)}
\label{sec:method}

\figFeatAndGroup{t}

We approach unsupervised semantic segmentation as an unsupervised pixel-wise feature learning problem (\fig{featAndGroup}).  The basic idea is that, once every pixel is transformed into a point in the feature space, image segmentation becomes a point clustering problem.  

Semantic segmentation and feature clustering form a pair of dual processes: {\bf 1)}
Clustering of feature $X$ defines segmentation $G$ in each image:  Pixels with features in the same (different) clusters belong to the same (different) semantic regions.   This idea is used to 
 co-segment similar images given handcrafted features~\cite{rother2006cosegmentation,joulin2010discriminativecoseg,lee2010collect}.
{\bf 2)}
Segmentation $G$ defines the similarity of feature $X$: A pixel should be mapped close to its own segment group and far from other segment groups in the feature space.  This idea is used to learn the pairwise feature similarity \cite{meila2000walk} and pixel-wise feature \cite{hwang2019segsort,ke2021spml} given segmentations.

Our key insight is that a good representation shall reveal not just a particular level of grouping -- as past co-segmentation methods have explored,
but any level of grouping in a consistent and predictable manner.  
If we embrace the ambiguity of grouping granularity that all previous methods have avoided and desire the consistency of hierarchical semantic segmentation on the pixel-wise feature, we address not only the shortcoming of cosegmentation, but also provide a joint feature-segmentation learning solution.

Specifically, while there is no supervision available for either feature $X$ or segmentation $G$, we can desire that: {\bf 1)} each segmentation separates features well and {\bf 2)} the coarser segmentation defined by next-level feature clusters simply {\it merges} the current finer segmentation.  These strong constraints guide the feature learning towards quality hierarchical segmentations, thereby better capturing semantics.

Our model has two components: {\bf 1)} 
multiview cosegmention to robustify feature clustering against spatial transformation and appearance variations of visual scenes, and {\bf 2)} clustering transformers to enforce consistent semantic segmentations across different levels of the feature grouping hierarchy.  Both are necessary for mapping pixel features to segmentations, which in turn impose desired pairwise attraction and repulsion on the pixel features.

In the following, we first introduce our contrastive feature learning loss given any grouping $G$, and then describe how we obtain three kinds of groupings within and across images, and how we evaluate their goodness of grouping and enforce their consistency.

\subsection{Pixel-Segment Contrastive Feature Learning}

We learn a pixel-wise feature extraction function $f$ as a convolutional neural network (CNN) with parameters $\theta$.  It transforms image $I$ to its pixel-wise feature $V$.  Let $\pmb{v}_i$ be the {\it unit-length} feature vector at pixel $i$ of image $I$:
\begin{align}
    \pmb{v}_i = f_i(I;\theta), \quad \|\pmb{v}_i\| = 1.
\end{align}
Suppose that $I$ is partitioned into segments (\fig{coseg}).  Let $\pmb{u}_s$ be the feature vector for segment $s$, defined as the (length-normalized) average pixel feature within the segment:
\begin{align}
    \pmb{u}_s \propto \text{mean}\left(\pmb{v}_i: i \text{ in segment } s\right), \quad \|\pmb{u}_s\| = 1
\end{align}

Consider a batch of images and their pixel groupings $\{(I,G)\}$.  We want to learn the right feature mapper $f$ such that all the pixels form distinctive clusters in the feature space, each corresponding to a different semantic group.

We follow \cite{hwang2019segsort,ke2021spml} to formulate desired feature-wise attraction and repulsion {\it not between pixels}, but {\it between pixels and segments}.  Such contrastive learning across granularity levels reduces computation, improves balance between attraction and repulsion, and is more effective \cite{wang2021unsupervised}.

Our contrastive feature learning loss to minimize is:
\begin{align}
\lossF(G)\!=\!\sum_{i}\!-\!\log\! 
\frac{\sum\limits_{s \in G_i^+}\! 
\exp{\frac{\pmb{v}_{i}^\top \pmb{u}_{s}}{T}}}
{\sum\limits_{s \in G_i^+}\! 
\exp{\frac{\pmb{v}_{i}^\top \pmb{u}_{s}}{T}} + 
\sum\limits_{s \in G_i^-}\! 
\exp{\frac{\pmb{v}_{i}^\top \pmb{u}_{s}}{T}}}
\label{eqn:lossF}
\end{align}
where $T$ is a temperature hyper-parameter that controls the concentration level of the feature distribution.  Ideally, $\pmb{v}_i$ should be attracted to segments in the positive set $G_i^+$ and repelled by segments in the negative set $G_i^-$.

Our batch of images consists of several augmented {\it views} of some training instances.  For pixel $i$ in a particular view of image $I$, $G_i^+$ includes segments of the same semantic group in any view of image $I$ except $i$'s own segment, in order to achieve within-instance invariance, whereas  $G_i^-$ includes segments of different semantic groups in any view of $I$, {\it and} segments of training instances other than $I$, in order to maximize between-instance discrimination \cite{wu2018unsupervised,hwang2019segsort}. 

\subsection{Consistent Segmentations by View \& Hierarchy}

\figCoseg{t}

From pixel feature $V$, we compute feature grouping $G_0$ and cluster feature $X_0$.  Our initial pixel grouping $G_e$ is based on OWT-UCM edges detected in the image.  Next-level cluster feature $X_{l+1}$ and grouping $G_{l+1}$ are predicted from $G_l$ with ensured consistency.  We use three levels for the sake of illustration (\fig{coseg}), but our procedure can be repeated for more (coarser) levels.

\noindent
{\bf Base cluster feature $X_0$ and grouping $G_0,G_e$}.
We segment each view of $I$ by clustering pixel features, resulting in base grouping $G_0$ and cluster (centroid) feature $X_0$ (\fig{featAndGroup}).

During training but {\it not} testing, we segment image $I$ into a fixed number of coherent regions according to its OWT-UCM edges \cite{dollar2014fast}, based on which
we split each $G_0$ region to obtain edge-conforming {\it segments} \cite{hwang2019segsort} marked by white lines in \fig{coseg}.
For training, we obtain pixel grouping $G_e$ by inferring the coherent region segmentation according to how each view is spatially transformed from $I$.

Minimizing $\lossF(G_e)$ encourages the feature to be similar not only for different pixels of similar appearances in the image, but also for corresponding pixels of different appearances across views of $I$.  The former grounds the feature $f$ at respecting low-level appearance coherence, whereas the latter develops view invariance in the feature.

\noindent
{\bf Next-level cluster feature $X_{l+1}$ and grouping $G_{l+1}$}.
Now we have grouping $G_0$ in the feature space of $V$, and for each cluster, we obtain its centroid feature in $X_0$.  We model how cluster feature $X_l$ maps to cluster feature $X_{l+1}$, which corresponds to how segmentation at level $l$ maps to segmentation at level $l+1$ in the image.

We adopt a probabilistic framework, where any feature point $\pmb{x}$ has a (soft assignment) probability belonging to a group determined by its cluster centroid.  Let $P_l(a)$ be the probability of $\pmb{x}$ in group $a$ at level $l$:
\begin{align}
P_{l}(a) = \text{Prob}(G_l=a \,|\,\pmb{x}).
\end{align}
To ensure that feature points in the same group remain together at the next level,  we introduce group transition probability $C_{l}^{l+1}(a,b)$, the transition probability from group $a$ at level $l$ to group $b$ at level $l\!+\!1$:
\begin{align}
C_{l}^{l+1}(a,b) & = \text{Prob}(G_{l+1}=b \,|\, G_{l}=a).
\end{align}
Per the Bayesian rule, we have:
\begin{align}
P_{l+1}(b) & = \sum_{a} P_l(a) \cdot C_{l}^{l+1}(a,b).
\label{eqn:hrchy_bayesian}
\end{align}
Writing $P_l$ as a row vector, we can derive the soft group assignment $P_{l+1}$ for cluster feature $X_0$ at level $l\!+\!1$:
\begin{align}
P_{l+1} = P_{l} \times C_l^{l+1} 
= P_0\times C_0^1 \times C_1^2 \times \cdots \times C_{l}^{l+1}.
\end{align}

\figClusterTransformer{t}
%\figHierTransformer{t}

\noindent
{\bf Clustering Transformers}.  $C_{l}^{l+1}$ is defined on multiview cosegmentation of each instance.  We learn a function, in terms of a transformer~\cite{carion2020end}, to naturally capture  feature group transitions for all the training instances.  It enables more consistent grouping compared to non-parametric clustering methods such as KMeans, NCut~\cite{von2007tutorial}, and FINCH~\cite{sarfraz2019efficient}.

Our clustering transformer from level $l$ to $l\!+\!1$ maps group centroid feature $X_l$ to the next-level group centroid feature $X_{l+1}$, and simultaneously outputs the group transition probability $C_l^{l+1}$ (\fig{clusterTransformer}).

\noindent
{\bf Consistent feature groupings. }
At level $l\!=\!0$, $P_0$ has binary values, indicating hard grouping $G_0$.  For next level $l$, we compute $P_{l+1}$ by propagating $P_l$ with our clustering transformer $C_{l}^{l+1}$, which also outputs $X_{l+1}$.
We obtain $G_{l+1}$ by binarizing $P_{l+1}$ with winner-take-all.  
By decreasing the number of groups as $l$ increases, we obtain consistent fine to coarse segmentations $G_1,G_2$ (Fig.~\ref{fig:featAndGroup}).  

Minimizing $\lossF(G_1)$ and $\lossF(G_2)$ encourages the feature $f$ to capture semantics at multiple levels and produce consistent hierarchical segmentations (\fig{coseg}).

\figFramework{!t}

\subsection{Goodness of Grouping}

While clustering transformers ensure grouping consistency across levels, we still need to drive feature learning towards good segmentations.  We follow  ~\cite{tsitsulin2020graph}
and supervise our transformer with modularity maximization ~\cite{newman2006finding} and collapse regularization.  The former seeks a partition that results higher (lower) in-cluster (out-cluster) similarity than the total expectation, whereas the latter encourages 
partitions of equal sizes.  We additionally maximize the separation between cluster centroids.

We first build a sparsified graph based on pairwise feature similarity for $X_0$.  Let $e$ be the number of edges in this graph, $n_l$ the number of centroids in $X_l$, $A$ the $n_0\!\times\! n_0$ connection matrix for edges, $D$ the $n_0\!\times\! 1$ degree vector of $A$, $M_l$ the $n_0\!\times\! n_l$ soft assignment matrix where each row is $P_l$ for a centroid of $X_0$, and $\pmb{z}_{l,k}$ the normalized $k$-th feature of $Z_l$ in \fig{clusterTransformer}.  Our goodness of grouping loss is:
{\small\begin{align}
\lossG &\!=\!\sum_{l\ge 1} \underbrace{\frac{-1}{2e}\text{trace}(M_l^\top (A\!-\!\frac{1}{2e} DD^\top) M_l)}_{\text{maximize modularity}}
\!+\! \underbrace{\frac{\sqrt{n_l}}{n_0} \|1^\top M_l \|_F\!-\! 1}_{ \text{collapse regularization}} \nonumber\\ 
&\!+\! \underbrace{\frac{1}{n_l}\sum_k -\log
\frac{\exp(\pmb{z}_{l,k}^\top \pmb{z}_{l,k})}
{\sum_j\exp(\pmb{z}_{l,k}^\top \pmb{z}_{l,j})} }_{\text{maximize centroid separation}}
\end{align}}

\subsection{Model Overview: Training and Testing}
\label{subsec:procedure}

Our model (\fig{framework}) is trained with the 
contrastive feature learning losses given edge-based grouping $G_e$ and multi-level feature-based grouping $G_l$, and the goodness of grouping loss, weighted by $\lambda_{E}$, $\lambda_{F}$, and $\lambda_{G}$ respectively:
\begin{align}
\mathcal{L}(f) =
\lambda_E \lossF(G_e)
+
\lambda_F \sum_{l\ge 1} \lossF(G_l)
+
\lambda_G L_g.
\end{align}
For testing, the same pipeline with the pixel feature CNN and clustering transformers predicts hierarchical segmentations $\{G_l\}$.
To benchmark segmentation performance given a labeled set, 
We follow \cite{hwang2019segsort} and predict the labels using 
k-nearest neighbor search for each segment feature.

\iffalse
{\small
\begin{align}
\mathcal{L}(f) \!=\! 
\underbrace{\lambda_E \lossF(G_e)}_{\text{coherent region}} 
+
\underbrace{\lambda_F \sum_{l\ge 1} \lossF(G_l)}_{\text{feature grouping}}
+
\underbrace{\lambda_G L_g}_{\text{goodness of grouping}}.
\end{align}
}
\fi

\def\tabCocoKittiCityscapes#1{
    \begin{table}[#1]
    \centering
    \resizebox{0.99\linewidth}{!}{%
    \begin{tabular}{l|c c|c c|c c}
    \toprule
    Training set & \multicolumn{2}{c|}{MSCOCO} & \multicolumn{2}{c|}{Cityscapes} & \multicolumn{2}{c}{KITTI-STEP}\\
    \midrule
    Validation set & \multicolumn{2}{c|}{VOC} & \multicolumn{2}{c|}{Cityscapes} & \multicolumn{2}{c}{KITTI-STEP}\\
    \midrule\midrule
    Method & mIoU & Acc. & mIoU & Acc. & mIoU & Acc.\\
    \midrule
    Moco~\cite{he2020momentum} & 28.1 & - & 15.3 & 69.5 & 13.7 & 60.3\\
    DenseCL~\cite{wang2021dense} & 35.1 & - & 12.7 & 64.2 & 9.3 & 47.6\\
    Revisit~\cite{van2021revisiting} & 35.1 & - & 17.1 & 71.7 & 17.0 & 65.0\\
    SegSort~\cite{hwang2019segsort} & 11.7 & 75.1 & 24.6 & 81.9 & 19.2 & 69.8\\
    Our \algorithmShort & \hl{41.9} & \hl{85.7} & \hl{32.5} & \hl{86.0} & \hl{21.7} & \hl{73.8}\\
    \bottomrule
    \end{tabular}}
    \caption{Our method delivers better performance on different types of datasets.  The results are reported on VOC, KITTI-STEP and Cityscapes val set, using IoU and pixel accuracy metrics.  In VOC, object categories are separated according to image scenes.  In Cityscapes and KITTI-STEP, images all come from urban street scene and thus contain mostly the same set of categories.  Instance-discrimination methods apply image-wise contrastive loss, and learn less optimally on Cityscapes and KITTI-STEP, as image scenes are similar.  Our \algorithmShort instead learns to discriminate regions at different scales and performs well on both types of datasets.}
    \label{tab:coco_kitti_cityscapes}
    \end{table}
}

\def\tabCocostuffPotsdam#1{
    \begin{table}[#1]
    \centering
    \resizebox{\linewidth}{!}{%
    \begin{tabular}{l|c c|c c}
    \toprule
     & \multicolumn{2}{c|}{COCO-stuff} & \multicolumn{2}{c}{Potsdam}\\
    \midrule
    Method & mIoU & Acc.& mIoU & Acc.\\
    \midrule
    %K-means~\cite{pedregosa2011scikit} & - & 14.1 & - & 35.3\\
    DeepCluster 2018~\cite{caron2018deep} & - & 19.9 & - & 29.2\\
    %SIFT~\cite{lowe2004distinctive} & - & 20.2 & - & 28.5\\
    Doersch 2015~\cite{doersch2015unsupervised} & - & 23.1 & - & 37.2\\
    Isola 2016~\cite{isola2015learning} & - & 24.3 & - & 44.9\\
    IIC~\cite{ji2019invariant} & - & 27.7 & - & 45.4\\
    AC~\cite{ouali2020Auto} & - & 30.8 & - & 49.3\\
    \midrule
    %SegSort (FCN)~\cite{hwang2019segsort} & 21.8 & 55.8 & 42.67 & 66.3\\
    SegSort~\cite{hwang2019segsort} & 16.4 & 49.9 & 35.0 & 59.0\\
    Our \algorithmShort & \hl{23.8} & \hl{57.6} & \hl{43.8} & \hl{67.4}\\
    \bottomrule
    \end{tabular}}
    \caption{Our method outperforms baselines on both stuff region and aerial scene parsing datasets.  The results are reported on COCO-stuff and Potsdam test set, using IoU and pixel accuracy metrics.  We evaluate our model using nearest neighbor search.  Our \algorithmShort achieves superior performance.}
    \label{tab:cocostuff_potsdam}
    \end{table}
}

\def\tabAblationLoss#1{
    \begin{table}[#1]
    \centering
    \resizebox{0.8\linewidth}{!}{%
    \begin{tabular}{c c c |c|c}
    \toprule
    $\lambda_{E}$ & $\lambda_{G}$ & $\lambda_{F}$ & single-view & multi-view\\
    \midrule
    %\rowcolor{Gray}
    %Single View & \checkmark & - & - & 9.7 & 74.2\\
    \checkmark & - & - & 13.0 & 40.9\\
    \checkmark & \checkmark & - & 13.8 & 41.7\\
    \checkmark & \checkmark & \checkmark & 14.0 & 41.9\\
    %\rowcolor{Gray}
    %Multiple View & \checkmark & - & - & 35.6 & 83.2\\
    %2 & \checkmark & - & - & 40.9\\
    %2 & \checkmark & \checkmark & - & 41.7\\
    %2 & \checkmark & \checkmark & \checkmark & 41.9\\
    \bottomrule
    \end{tabular}}
    \caption{Regularizing with our goodness of grouping loss and pixel-to-segment contrastive losses improves learned features.  The results are reported over VOC val set, using IoU metric. Our resulted pixel features encode better semantic information.}
    \label{tab:ablation_loss}
    \end{table}
}

\def\tabAblationClustering#1{
    \begin{table}[#1]
    \centering
    \resizebox{0.99\linewidth}{!}{%
    \begin{tabular}{l|c c c c}
    \toprule
    Method & KMeans & NCut~\cite{von2007tutorial} & FINCH~\cite{sarfraz2019efficient} & Our Transfomer\\
    \midrule
    mIoU & 41.2 & 41.3 & 40.6 & \hl{41.9}\\
    \bottomrule
    \end{tabular}}
    \caption{Our hierarchical clustering transformer follows semantics closer than other non-parametric clustering algorithms.  The results are reported on VOC val set with IoU metric.  Our learned representations achieve better unsupervised semantic segmentation.}
    \label{tab:ablation_clustering}
    \end{table}
}

\def\rowVS#1{
\imw{img_#1}{0.249}&
\imw{segsort_#1}{0.249}&
\imw{hsg_#1}{0.249}&
\imw{gt_#1}{0.249}\\[-2pt]
}
\def\figVisSemantic#1{
    \begin{figure}[#1]
        \centering
        \tb{@{}cccc@{}}{0.1}{
        \rowVS{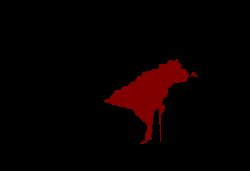}
        \rowVS{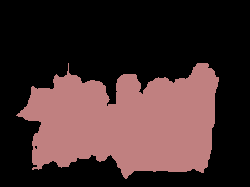}
        %\rowV{2007_008415}
        \rowVS{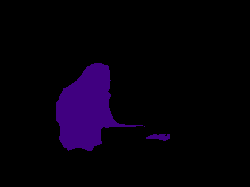}
        \midrule
        \rowVS{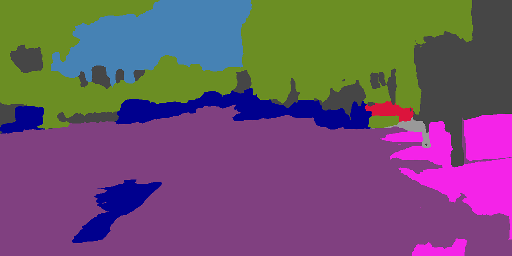}
        %\rowV{frankfurt_000001_025512}
        \rowVS{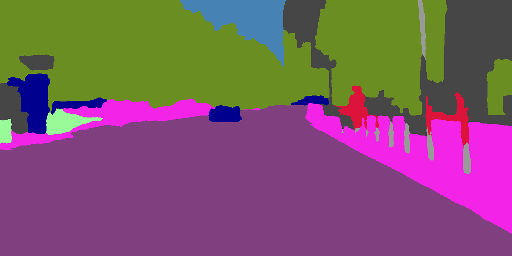}
        \rowVS{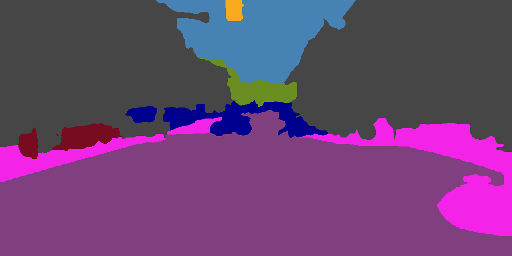}
        \midrule
        %\rowV{0007_000134}
        \rowVS{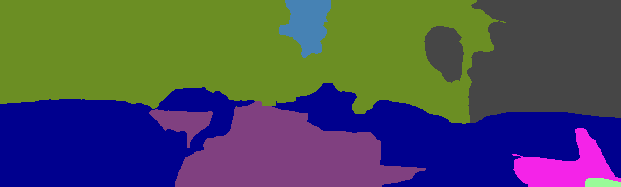}
        %\rowV{0007_000590}
        %\rowV{0007_000743}
        %\rowV{0008_000199}
        \rowVS{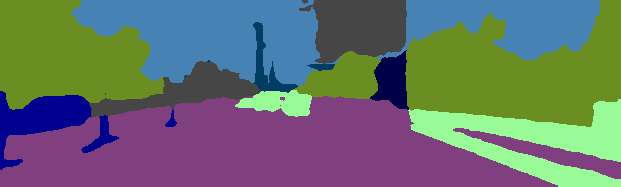}
        \rowVS{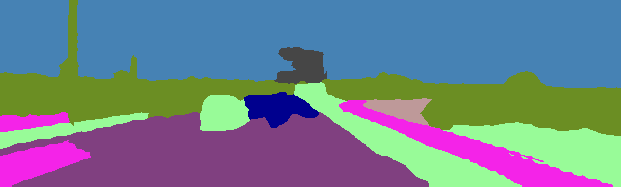}
        %\rowV{0010_000179}
        image&
        SegSort&
        \algorithmShort&
        ground truth\\
        }
        \caption{Our framework performs better on different types of datasets.  From top to bottom every three rows are visual results from VOC, Cityscapes and KITTI-STEP dataset.  The results are predicted via segment retrievals.  Our pixel-wise features encode more precise semantic information than baselines.}
        \label{fig:vis_semantic}
    \end{figure}
}

\def\rowVH#1{
\tb{@{}cccc@{}}{0.1}{%
\imwjpg{img_#1}{0.249}&
\imw{tf1_#1}{0.249}&
\imw{tf3_#1}{0.249}&
\imw{tf4_#1}{0.249}\\[-2pt]
\imw{edge_#1}{0.249}&
\imw{se1_#1}{0.249}&
\imw{se3_#1}{0.249}&
\imw{se4_#1}{0.249}\\
}}
\def\rowFVH#1{
\tb{@{}cccc@{}}{0.1}{%
image & 12 regions & 6 regions & 3 regions\\
\imwjpg{img_#1}{0.249}&
\imw{tf1_#1}{0.249}&
\imw{tf3_#1}{0.249}&
\imw{tf4_#1}{0.249}\\[-2pt]
\imw{edge_#1}{0.249}&
\imw{se1_#1}{0.249}&
\imw{se3_#1}{0.249}&
\imw{se4_#1}{0.249}\\
}}
\def\figVisHierarchy#1{
    \begin{figure}[#1]
        \centering
        \tb{c}{0}{%
        \imw{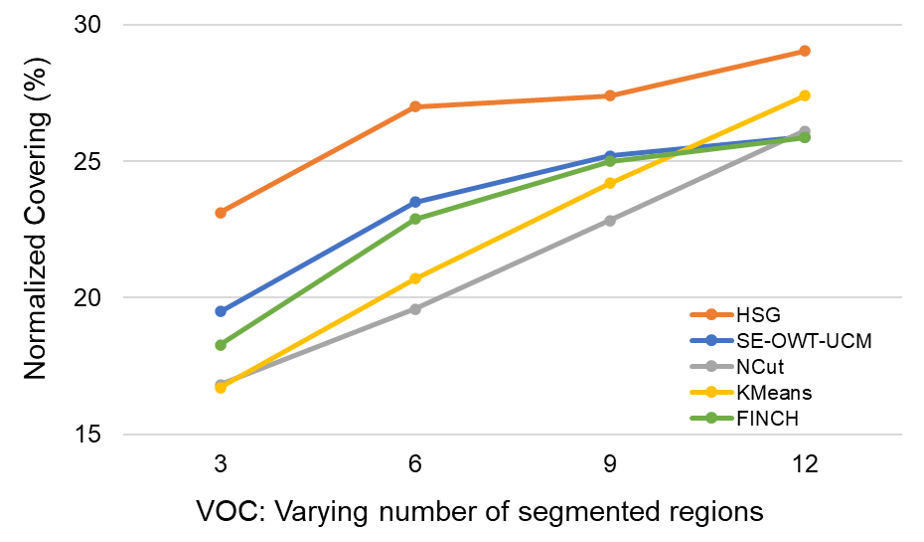}{1.0}\\[5pt]
        \midrule
        %\rowFVH{2007_000629}\\
        \rowFVH{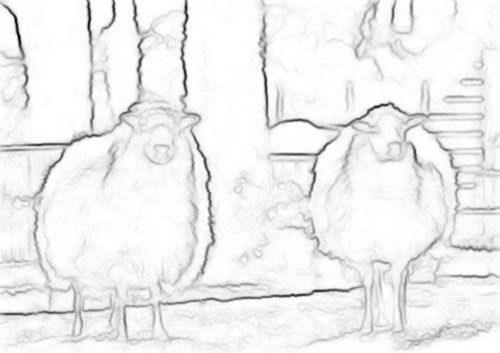}\\
        \rowVH{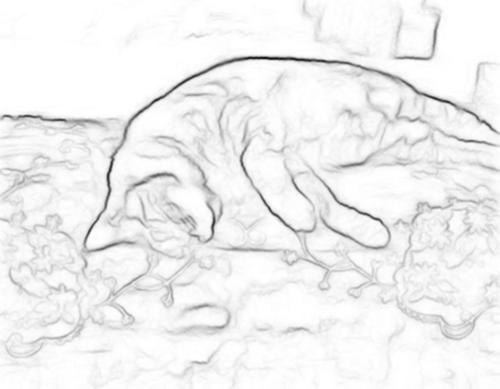}\\
        %\rowVH{2007_004483}\\
        %\rowVH{2007_009245}\\
        \rowVH{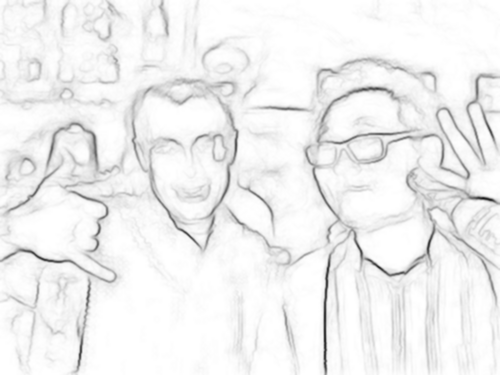}\\
        \rowVH{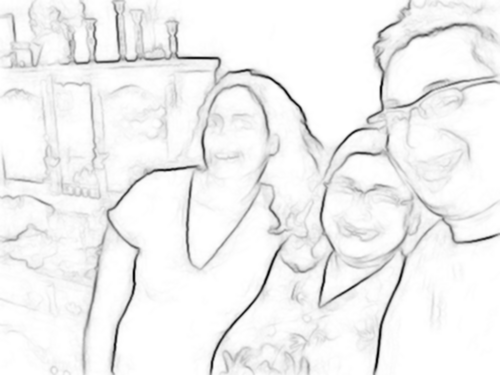}\\
        %\rowVH{2009_002185}\\
        %\rowVH{2009_003666}\\
        }
        \caption{Our clustering transformers capture semantics at different levels of granularity.  {\bf Top:} We compare to other clustering algorithms on VOC val set, using {\it Normalized Foreground Coverings} as metric.  We exclude background regions for evaluation.  Our \algorithmShort overlaps with ground truths more accurately.  {\bf Bottom:} We present visual results to compare our hierarchical segmentation (top row) with SE~\cite{dollar2014fast}-OWT-UCM procedure (bottom row).  We also show the detected edges at the leftmost figure in the bottom row.   Each image is segmented into $12$, $6$, $3$ regions.   Our method reveals low-to-high level of semantics more consistently.}
        \label{fig:vis_hierarchy}
    \end{figure}
}

\section{Experiments}
\label{sec:exp}

We benchmark our model on two tasks: unsupervised semantic segmentation and hierarchical image segmentation, the first on five major object- and scene-centric datasets and the second on Pascal VOC.  We conduct ablation study to understand the contributions of our model components.

We adopt FCN-ResNet50 as the common backbone architecture.  The FCN head consists of $1 \times 1$ convolution, $\mathrm{Batch Norm}$, $\mathrm{ReLU}$, and $1 \times 1$ convolution.  Specifically, we follow DeepLabv3~\cite{chen2017rethinking} to set up the dilation and strides in ResNet50.  We set $\mathrm{Multi\_Grid}$ to $(1,2,4)$ in $\mathrm{res5}$.  The $\mathrm{output\_stride}$ is set to $16$ and $8$ during training and testing.    We do not use any pre-trained models, but train our models from scratch on each dataset.
Ground-truth annotations are not for training but only for testing and evaluation's sake.

\noindent\textbf{Pascal VOC 2012}~\cite{everingham2010pascal} is a generic semantic segmentation dataset of 20 object category and a background class. It consists of $1,464$ and $1,449$ images for training and validation.  We follow  ~\cite{chen2016deeplab} to augment the training data with additional annotations~\cite{hariharan2011semantic}, resulting in $10,582$ training images.  Following ~\cite{van2021revisiting}, we do not train but only inference on VOC.

\noindent\textbf{MSCOCO}~\cite{lin2014microsoft} is a complex scene parsing dataset with $80$ object categories. Objects are embedded in more complex scenes, with more objects per image than Pascal ($7.3$ vs. $2.3$).  Following ~\cite{wang2021dense,van2021revisiting}, we use {\it train2017} split ($118,287$ images) for training and test on the VOC validation set.

\noindent\textbf{Cityscapes}~\cite{cordts2016cityscapes} is an urban street scene parsing dataset,  with $19$ stuff and object categories.  Unlike MSCOCO and VOC where classes are split by scene context, Cityscapes contains similar street scenes covering  almost all $19$ categories.  
The train/test split is $2,975/500$.

\noindent\textbf{KITTI-STEP}~\cite{weber2021step} is a video dataset for urban scene understanding, instance detection and object tracking.  It has pixel-wise labels of the same $19$ categories as Cityscapes.  There are $12$ and $9$ video sequences for training and validation, or $5,027$ and $2,981$ frames.

\noindent\textbf{COCO-stuff}~\cite{caesar2018coco} is a scene texture segmentation dataset,  a subset of MSCOCO. As ~\cite{ji2019invariant,ouali2020Auto},  we use 15 coarse {\it stuff} categories and reduce the dataset to 52K images with at least $75\%$ stuff pixels.  The train/test split is $49,629 / 2,175$.

\noindent\textbf{Potsdam}~\cite{gerke2014use} is a dataset for aerial scene parsing. The raw $6000 \!\times\! 6000$ image is divided into $8550$ RGBIR $200 \!\times\! 200$ patches.  There are 6 categories ({\it roads, cars, vegetation, trees, buildings, clutter}). The train/test split is $7,695/855$.

\tabCocoKittiCityscapes{h}
\tabCocostuffPotsdam{h}
\tabAblationLoss{h}
\tabAblationClustering{h}
\figVisSemantic{h}
\figVisHierarchy{!tp}
%\figMetricHierarchy{h}

% \subsection{Main Results and Ablation Study}
% \label{subsec:main_results}

\noindent\textbf{Results on unsupervised semantic segmentation.}
All the models are trained from scratch and evaluated by IoU and pixel accuracy. For VOC, we follow baselines~\cite{van2021revisiting} to train on MSCOCO.  Table.~\ref{tab:coco_kitti_cityscapes} shows that our method outperforms baselines by $6.8\%$, $7.9\%$ and $2.5\%$ in mIoU on VOC, Cityscapes, and KITTI-STEP validation sets respectively.

Note that methods relying on image-wise instance discrimination do not work well on Cityscapes and KITTI-STEP.  Both datasets have urban street scenes with similar categories in each image.  Our method can still discover semantics by discriminating regions among these images.

For texture segmentation on COCO-stuff and Potsdam,   Tab.~\ref{tab:cocostuff_potsdam} shows that our method achieves huge gains, $+26.8\%$ and $+18.1\%$  over IIC~\cite{ji2019invariant} and AC~\cite{ouali2020Auto} respectively.

\noindent\textbf{Results on hierarchical segmentation.}  We benchmark hierarchical segmentation with respect to ground-truth segmentation.  We evaluate the overlapping of regions between predicted segmentations and ground truth within each image, known as {\it Segmentation Covering}~\cite{arbelaez2010contour}.  
However, such a metric scores performance with the number of pixels within each segment, and is thus easily biased towards large regions.  For  object-centric dataset VOC, a trivial all-foreground mask would rank high by the Covering metric.

We propose a {\it Normalized Foreground Covering} metric, by focusing on the foreground region and the overlap ratio instead of the overlap pixel count.
To measure the average foreground region overlap ratio of a ground-truth segmentation $S$ by a predicted segmentation $S'$, we define: 
{\small\begin{align}
\mathrm{NFCovering}(S' \!\rightarrow\! S_{fg}) \!=\! \frac{1}{|S_{fg}|} \sum\limits_{R \in S_{fg}} \max\limits_{R' \in S'} \frac{|R \cap R'|}{|R \cup R'|}
\end{align}}
where $S_{fg}$ denotes the set of ground-truth foreground regions.  Given a hierarchical segmentation, we report NFCovering at each level in the hierarchy.  Fig.~\ref{fig:vis_hierarchy} shows that our clustering transformers produce segmentations better aligned with the ground-truth foreground at every level.

% \subsection{Visualization}
% \label{subsec:visualization}
\noindent\textbf{Visualization.}  Fig.~\ref{fig:vis_semantic} shows sample semantic segmentations on VOC (trained on MSCOCO), Cityscapes and KITTI-STEP.  Compared to SegSort~\cite{hwang2019segsort}, our method retrieves same-category segments more accurately.  For larger objects or stuff categories, such as {\it airplane} or {\it road}, our results are more consistent within the region.  Our segmentations are also better at respecting object boundaries.

We also compare our hierarchical segmentations with SE~\cite{dollar2014fast}-OWT-UCM, an alternative based entirely on low-level cues.  Fig.~\ref{fig:vis_hierarchy} bottom shows that, when partitioning an image into $12$, $6$ and $3$ regions, our segmentations follow the semantic hierarchy more closely.

\noindent\textbf{Ablation study.}
Tab.~\ref{tab:ablation_loss} shows that 
our model improves consistently by adding the feature learning loss based on hierarchical groupings and the goodness of grouping loss.   It also shows that multiview cosegmentation significantly improves the performance over a single image.

Tab.~\ref{tab:ablation_clustering} shows that our clustering transformers provide better regularization with hierarchical groupings than alternative  non-parametric clustering methods.  

% We deliver the first unsupervised hierarchical semantic segmentation method by {\bf 1)} imposing consistent grouping hierarchy in the feature space with clustering transformers and {\bf 2)} imposing spatial consistency of grouping with multiview cosegmentation.  We formulate the task as a pixel-wise representation learning problem, optimizing the feature with the feature induced groupings at multiple levels of granularity.  
%  We demonstrate that our feature-induced hierarchical segmentations follow semantics more precisely and our unsupervised segmentation outperforms baselines on major object-centric and scene-centric benchmarks.
\noindent\textbf{Summary.}  We deliver the first unsupervised hierarchical semantic segmentation method based on multiview cosegmentation and clustering transformers.  
Our unsupervised segmentation outperforms baselines on major object- and scene-centric benchmarks, and our hierarchical segmentation discovers semantics far more accurately.

\vspace{5pt}

\noindent{\bf Acknowledgements.} This work was supported, in part, by Berkeley Deep Drive, Berkeley AI Research Commons with Facebook, NSF 2131111, and a Bosch research gift.

\clearpage
{\small
\bibliographystyle{ieee_fullname}
\bibliography{seg}
}

\clearpage
\maketitle
%%%%%%%%% Define tables.
\def\tabVoc#1{
    \begin{table*}[#1]
    \centering
    \resizebox{\linewidth}{!}{%
    \begin{tabular}{l|c c c c c c c c c c c c c c c c c c c c|c}
    \Xhline{1pt}
    Method & aero & bike & bird & boat & bottle & bus & car & cat & chair & cow & table & dog & horse & mbike & person & plant & sheep & sofa & train & tv & mIoU\\
    \hline\hline
    MaskContrast~\cite{van2021unsupervised} & {\bf 76.2} & 26.9 & 70.2 & 49.6 & 56.1 & 80.3 & 66.8 & 66.8 & 10.6 & 55.1 & 17.5 & 65.2 & 51.8 & 59.7 & 58.8 & 23.1 & {\bf 73.5} & 24.9 & 70.9 & 38.9 & 53.9\\
    SegSort~\cite{hwang2019segsort} & 71.3 & 26.4 & 70.7 & 56.1 & 51.9 & 78.2 & 68.5 & 72.1 & 12.7 & 47.2 & 36.4 & 65.3 & 45.5 & 61.6 & 63.7 & 29.3 & 60.0 & 30.1 & 70.3 & 59.4 & 55.5\\
    Our \algorithmShort & 75.1 & {\bf 32.2} & {\bf 76.9} & {\bf 60.4} & {\bf 63.9} & {\bf 81.7} & {\bf 75.5} & {\bf 82.0} & {\bf 18.5} & {\bf 48.7} & {\bf 51.2} & {\bf 71.5} & {\bf 55.0} & {\bf 69.4} & {\bf 71.0} & {\bf 39.8} & 66.8 & {\bf 33.3} & {\bf 72.3} & {\bf 59.6} & {\bf 61.7}\\
    \hline
    {\it vs. baseline} & -1.1 & +5.3 & +6.2 & +4.3 & +7.8 & +1.4 & +7.0 & + 9.9 & +5.8 & +1.5 & +14.8 & +6.2 & +3.2 & +7.8 & +7.3 & +10.6 & -6.7 & +3.2 & +1.4 & +0.2 & +6.2\\
    \Xhline{1pt}
    \end{tabular}}
    \caption{Our method outperforms SegSort~\cite{hwang2019segsort} and MaskContrast~\cite{van2021unsupervised} among most categories on VOC val dataset.  We adopt PSPNet~\cite{zhao2016pyramid}-ResNet50 as our backbone CNN.  All models are supervisedly pre-trained on ImageNet~\cite{deng2009imagenet}.  Per-category performance is evaluated using IoU metric.  We use the officially trained model and released code for inference MaskContrast. Our \algorithmShort demonstrates superior performance for semantic segmentation.}
    \label{tab:supp_voc}
    \end{table*}
}

%%%%%%%%% Define figures.
\def\rowVS#1{
\imwh{img_#1}{0.24}{0.11}&
\imwh{segsort_#1}{0.24}{0.11}&
\imwh{hsg_#1}{0.24}{0.11}&
\imwh{gt_#1}{0.24}{0.11}\\[-2pt]
}
\def\figVisSemantic#1{
    \begin{figure}[#1]
        \centering
        \vspace{-10pt}
        \tb{@{}cccc@{}}{0.1}{
        \rowVS{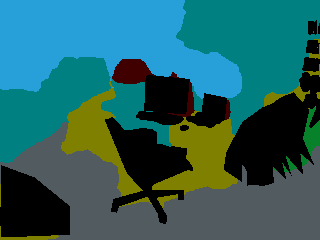}
        \rowVS{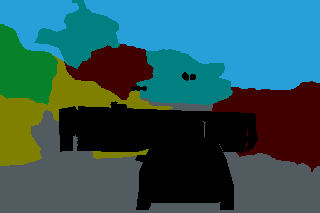}
        \rowVS{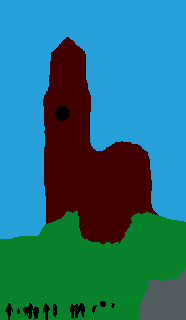}
        \midrule
        \rowVS{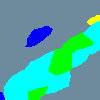}
        \rowVS{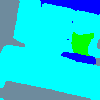}
        \rowVS{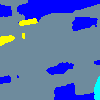}
        image&
        SegSort&
        \algorithmShort&
        ground truth\\
        }
        %\vspace{-5pt}
        \caption{Our framework delivers better semantic segmentation on textural and aerial region parsing datasets.  From top to bottom every three rows are visual results from COCO-stuff and Potsdam.  The results are predicted via segment retrievals.  Our results are more consistent and accurate within each category.  Additionally, our segmentation predictions are better aligned with the boundaries.  Our pixel-wise features encode more precise semantic information than baselines.}
        \label{fig:supp_vis_semantic}
    \end{figure}
}

\def\rowVC#1{
\imwh{#1Q}{0.199}{0.13}&
\imwh{#1R1}{0.199}{0.13}&
\imwh{#1R2}{0.199}{0.13}&
\imwh{#1R3}{0.199}{0.13}&
\imwh{#1R4}{0.199}{0.13}\\[-2pt]
}
\def\figVisContext#1{
    \begin{figure*}[#1]
        \centering
        \vspace{-10pt}
        \tb{@{}c|cccc@{}}{0.1}{
        \rowVC{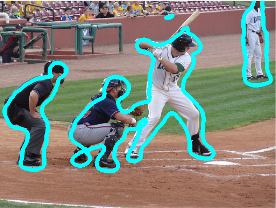}
        \rowVC{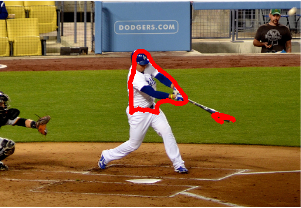}
        \rowVC{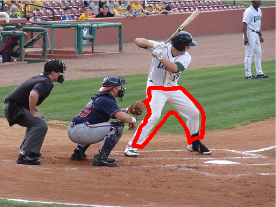}
        \midrule
        \rowVC{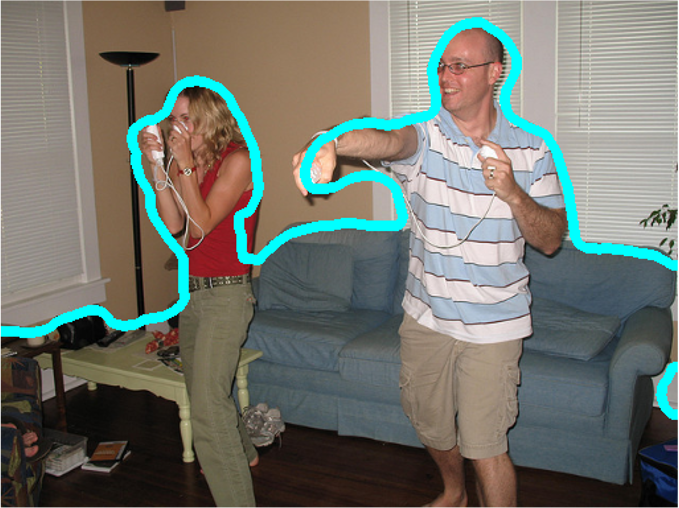}
        \rowVC{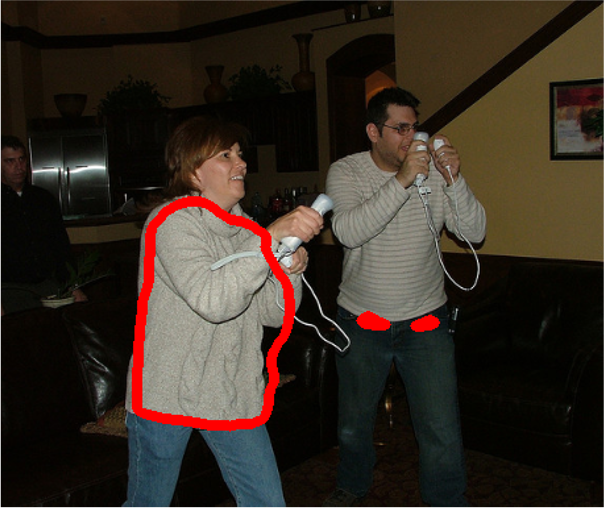}
        \rowVC{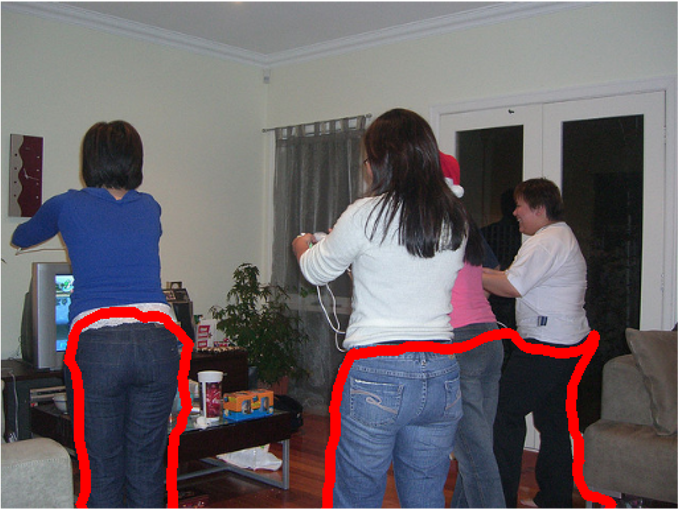}
        }
        %\vspace{-5pt}
        \caption{Sample retrieval results in MSCOCO for two images, baseball (Rows 1-3) and wii sport (Rows 4-6), based on our CNN features.
        Column 1 shows a query segment and Columns 2-5 are its nearest neighbour retrievals at the same level of the hierarchy.  Segments at a \textcolor{cyan}{coarser} / \textcolor{red}{finer} level are shown in cyan (Rows 1,4) / red (Rows 2-3, 5-6).
        Coarser segment retrievals show that our feature learned from hierarchical groupings are reflective of the visual scene layout (For example, Row 1 all has the 3-person baseball pitching configuration despite drastic appearance variations), whereas finer segment retrievals show that our learned feature is precise at characterizing both the segment itself and the visual context around it (For example, the feature of the query segment ({\it legs}) in Row 3 is indicative of the pitcher pose on the baseball field).  Such a holistic yet discriminative feature representation is discovered in a pure data-driven fashion without any semantic supervision.}
        \label{fig:supp_vis_context}
    \end{figure*}
}

\def\figTsne#1{
    \begin{figure*}[#1]
        \centering
        %\vspace{-5pt}
        \imw{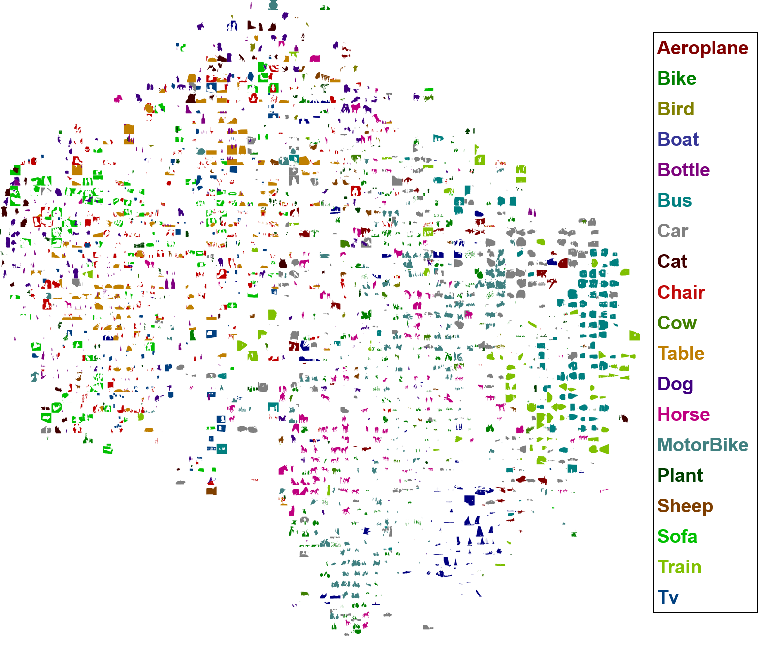}{1.0}
        %\vspace{-5pt}
        \caption{Our visual representations encode contextual information of co-occurring objects.  We visualize the average feature of {\it person} category region on Pascal VOC 2012 dataset using tSNE~\cite{maaten2008visualizing}.  We use the feature mappings extracted with models trained from scratch on MSCOCO.  We represent each {\it person} category region with the co-occurring object categories, and observe that features in the similar semantic context are clustered.}
        \label{fig:supp_tsne_context}
    \end{figure*}
}

\def\rowVA#1{
\imw{#1_image}{0.099}&
\imw{#1_finehrchy}{0.099}&
\imw{#1_fineattn0}{0.099}&
\imw{#1_fineattn1}{0.099}&
\imw{#1_fineattn2}{0.099}&
\imw{#1_fineattn3}{0.099}&
\imw{#1_fineattn4}{0.099}&
\imw{#1_fineattn5}{0.099}&
\imw{#1_fineattn6}{0.099}&
\imw{#1_fineattn7}{0.099}\\[-2pt]
}
\def\figVisAttention#1{
    \begin{figure*}[#1]
        \centering
        \vspace{-10pt}
        \tb{@{}cccccccccc@{}}{0.1}{
        %\rowVA{2007_007948}
        \rowVA{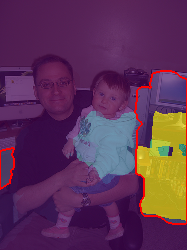}
        \rowVA{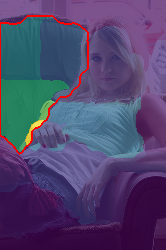}
        %\rowVA{2008_004892}
        \rowVA{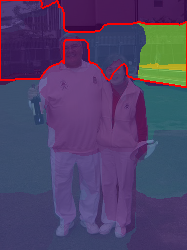}
        %\rowVA{2008_008212}
        %\rowVA{2009_000072}
        %\rowVA{2010_001140}
        %\rowVA{2010_001287}
        \rowVA{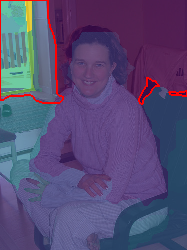}
        }
        \caption{The multi-head attention maps reveal the fine-to-coarse semantic relationships among image segments.  {\bf From left to right:} input image, our feature-induced segmentation, attention maps in the decoder of our clustering transformers.  We use a clustering transformer to partition each image into $8$ clusters, and show the attention map (colored in {\it viridis} color maps) of \textcolor{red}{each cluster} to all the image segments.  We observe these clusters correlate better with image segments that carry more similar semantic meanings, e.g., the `head' cluster attends more to body parts than background regions.  Such correlation information implies the next-level groupings: `head' will be grouped with `torso' instead of `background'}
        \label{fig:supp_vis_attention}
    \end{figure*}
}

\def\tabSpeed#1{
    \begin{table}[#1]
    \centering
    \resizebox{\linewidth}{!}{%
    \begin{tabular}{l|c c c c c}
    \toprule
     & No Hierarchy & \algorithmShort & KMeans & NCut & FINCH\\
    \midrule
    ms & 120 & 158 & 165 & 170 & 381\\
    \bottomrule
    \end{tabular}}
    \caption{Our method imposes less runtime overhead than other hierarchical clustering methods.  All methods are conducted on a $640 \times 640$ image, which is hierarchically partitioned into $25, 16, 9$ and $4$ segments.  While major latency comes from the pixel embedding network, HSG is still $17\%$ faster than KMeans.}
    \label{tab:speed}
    \end{table}
}

\def\rowO#1{
\tb{@{}c@{}}{0.1}{%
\imw{#1}{0.85}\\
(a)\\[-2pt]
}}
\def\rowD#1{
\tb{@{}cc@{}}{0.1}{%
\imw{part_#1}{0.5}&
\imw{person_#1}{0.5}\\
(b) & (c)\\[-2pt]
}}
\def\figDenseposeCompare#1{
    \begin{figure}[#1]
        \centering
        \vspace{-10pt}
        \tb{c}{0}{%
        \rowO{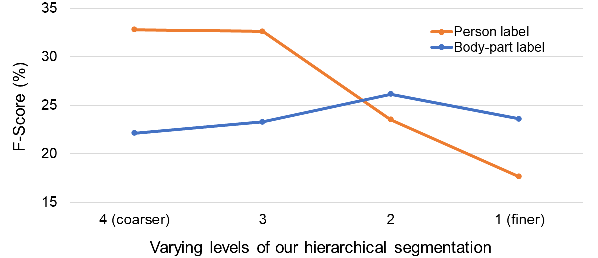}\\
        \rowD{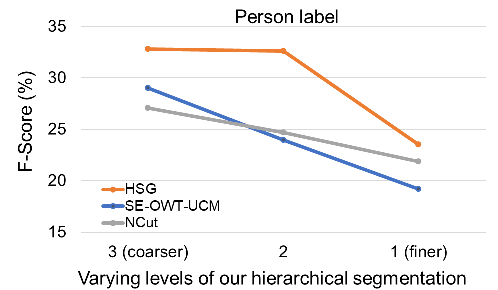}\\
        }
        %\vspace{-5pt}
        \caption{Our hierarchical segmentations outperform others, which better pick out semantics at different levels of granularity.  On DensePose~\cite{alp2018densepose}, ground-truth labels are processed at two levels of semantics: person and body parts.  {\bf a)} F-score of region matching among ground truths and our hierarchical segmentations, {\bf b)} F-score on fine (body-part-level) labels and {\bf c)} F-score on coarse (person-level) labels.  Hierarchical segmentation is needed to capture semantics across different granularity and our HSG outperforms others by large margin at every level.}
        \label{fig:densepose_fscore}
    \end{figure}
}

\def\figHrchyTransformerArch#1{
    \begin{figure}[#1]
        \centering
        \vspace{-20pt}
        \includegraphics[width=\linewidth]{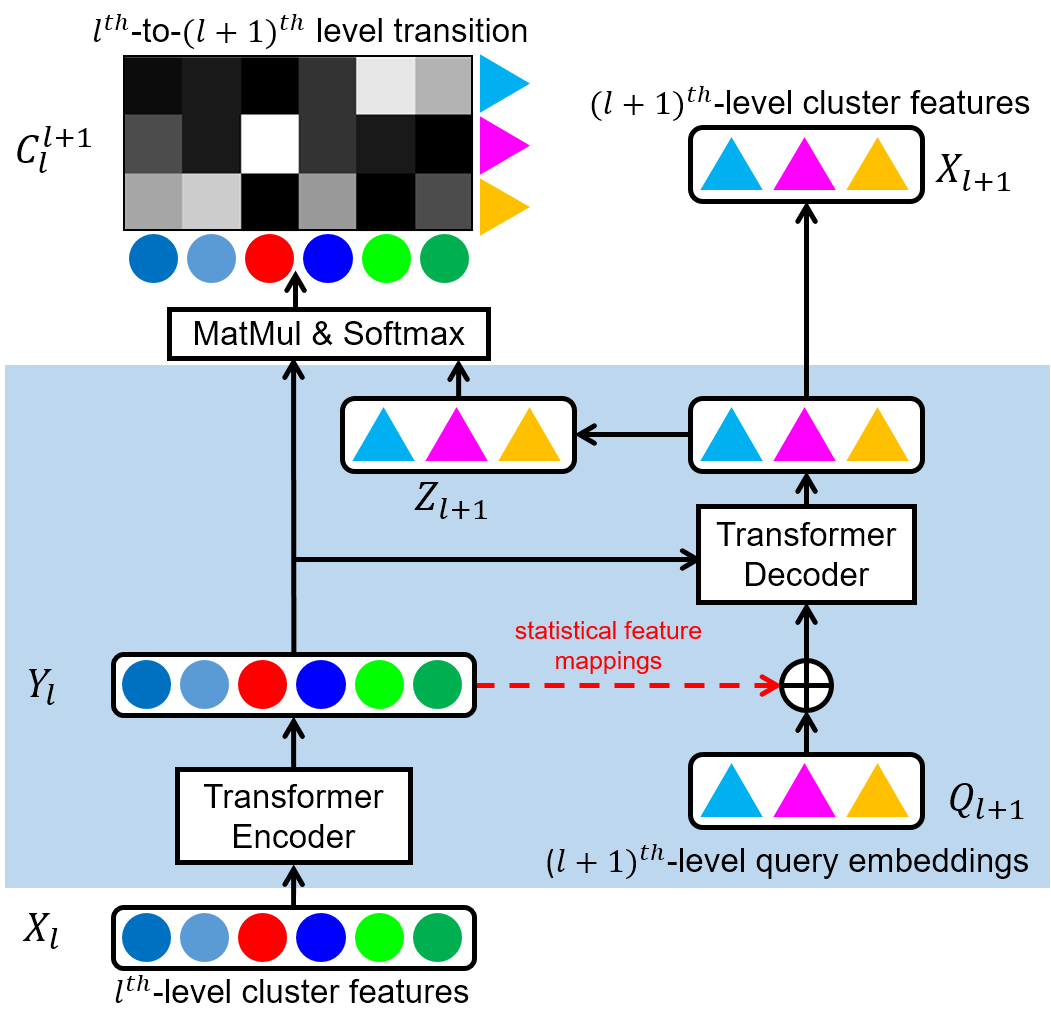}
        \caption{Our clustering transformer
        enforces grouping consistency across levels by mapping feature $X_l$ to $X_{l+1}$ with feature transition $C_l^{l+1}$.  $X_{l+1}$ and $C_l^{l+1}$ are learned simultaneously.
       Shown here for level $l\!=\!0$ in \fig{featAndGroup},
        the transformer encoder takes inputs $X_{l}$
         and outputs contextualized feature $Y_l$.
         The transformer decoder takes learnable inputs from query embeddings $Q_{l+1}$, and outputs $X_{l+1}$ and additionally projected feature $Z_{l+1}$.
         The transition is predicted as: $C_l^{l+1}\!=\!\text{softmax}\left(\frac{1}{\sqrt{m}}Y_l^\top Z_{l+1}\right)$; $m$ is the feature dimension.  {\color{red} Statistical feature mapping}:
         Calculate $Y_l$'s mean and std,  transform them by fc layers, and add to $Q_{l+1}$ for instance adaptation. }
        \label{fig:supp_hrchy_transformer}
    \end{figure}
}

\def\tabHyperParameters#1{
    \begin{table}[#1]
    \centering
    \resizebox{\linewidth}{!}{%
    \begin{tabular}{l|c c c c c|c c c}
    \toprule
    Dataset & B.S & C.S & L.R & W.D & Epochs & $\lambda_{E}$ & $\lambda_{G}$ & $\lambda_{F}$\\
    \midrule
    \rowcolor{Gray}
    MSCOCO & 128 & 224 & 0.1 & 0.0001 & 380 & 1.0 & 0.0 & 0.0\\
     & 48 & 448 & 0.008 & 0.0001 & 8 & 1.0 & 1.0 & 0.1\\
    %VOC & 16 & 448 & 0.002 & 0.0005 & 15 & 1.0 & 1.0 & 0.1\\
    Cityscapes & 32 & 448 & 0.1 & 0.0001 & 400 & 1.0 & 0.2 & 0.1\\
    KITTI-STEP & 48 & 448 & 0.1 & 0.0001 & 400 & 1.0 & 0.2 & 0.1\\
    COCO-stuff & 8 & 336 & 0.003 & 0.0005 & 5 & 1.0 & 0.2 & 0.1\\
    Potsdam & 8 & 200 & 0.003 & 0.0005 & 30 & 1.0 & 0.2 & 0.1\\
    \bottomrule
    \end{tabular}}
    \caption{Hyper-parameters for training on different datasets. Gray colored background indicates pre-training settings. B.S, C.S, L.R, W.D denote batch size, crop size, learning rate and weight decay.}
    \label{tab:supp_hyper_parameters}
    \end{table}
}

%%%%%%%%% BODY TEXT - ENTER YOUR RESPONSE BELOW
\section{Supplementary}
\label{sec:supp}

\figVisAttention{t!}
\figDenseposeCompare{h}
\figVisContext{t!}
\figTsne{t!}
\figVisSemantic{h}
\tabVoc{h}
\tabSpeed{h}
\figHrchyTransformerArch{h}
\tabHyperParameters{h}

We propose the first unsupervised hierarchical semantic segmentation method.  Our core idea has two folds: {\bf 1)} imposing grouping hierarchy in the feature space with clustering transformers and {\bf 2)} enforcing spatial grouping consistency with multiview cosegmentation.  We demonstrate state-of-the-art performance on unsupervised semantic segmentation and hierarchical segmentation.  In this supplementary, we include more details on the following aspects:

\begin{itemize}
    \item We present the visual results of the attention maps from our clustering transformers in~\ref{subsec:supp_attn}.
    
    \item We present the qualitative results of hierarchical semantic segmentation on DensePose dataset in~\ref{subsec:supp_hrchy_seg}.
    
    \item We present the visual results of contextual information encoded in our learned feature mappings in~\ref{subsec:supp_vis_context}.

    \item We showcase unsupervised semantic segmentation on COCO-stuff and Potsdam datasets in~\ref{subsec:supp_vis_semantic}.

    \item We present per-category results on VOC with ImageNet-trained models in~\ref{subsec:supp_unsup}.
    
    \item We present inference latency of our clustering transformers in~\ref{subsec:supp_latency}
    
    \item We describe more details of our clustering transformers in~\ref{subsec:supp_hrchy_transformer}.
    
    \item We describe the details of our experimental settings and hyper-parameters in~\ref{subsec:supp_exp_settings}
    
    \item We detail the experimental settings on VOC using ImageNet-trained models in~\ref{subsec:supp_imagenet_exp}.
\end{itemize}

% \section{Experimental Results}
% \label{sec:supp_results}

\subsection{Visual Results on Attention Maps from Decoder}
\label{subsec:supp_attn}
We visualize the multi-head attention maps in the decoder of our clustering transformer.  Such attention maps correspond to the correlation among cluster centroids and input segments. As shown in Fig.~\ref{fig:supp_vis_attention}, we observe that each cluster attends to their cluster members, e.g. face and hair of the head region.  Interestingly, we also see these clusters correlate better with image segments that carry more similar semantic meanings.  For example, the `head' cluster attends more to body parts than background regions.  Such correlation information implies the next-level groupings: `head' will be grouped with `torso' instead of `background'.

\subsection{Hierarchical Semantic Segmentation}
\label{subsec:supp_hrchy_seg}

We show that hierarchical segmentation is needed to pick out semantics at different levels of granularity.  On DensePose~\cite{alp2018densepose} dataset, we process ground-truth labels at two levels of semantics: person and body parts.  Body-part-level labels include head, torso, upper and lower limb regions.  As shown in the top of Fig.~\ref{fig:densepose_fscore}, coarse (person-level) and fine (body-part-level) labels are best picked out with higher and lower levels of segmentations.  The evaluation metric is based on F-score of region matching.

We next demonstrate the efficacy of our hierarchical clustering transformer.  As shown in the bottom of Fig.~\ref{fig:densepose_fscore}, our predicted segmentations outperform others by large margin at every level in the hierarchy on both the fine and coarse sets of semantic labels.  Hierarchical consistency provides regularizations which help us to obtain better segmentation results at any granularity.

\subsection{Visual Results on Contextual Retrievals}
\label{subsec:supp_vis_context}

 We reveal the encoding of visual context in our learned feature representations.  We first conduct hierarchical segmentation using our clustering transformers to partition an image into fine and coarse regions.  We then compute the unit-length average feature within each region and perform nearest neighbor search among the training dataset.  Fig.~\ref{fig:supp_vis_context} shows nearest neighbor retrievals at coarse (cyan) and fine (red) segmentations. The query and retrieved segments are generated at same level of partitioning. Strikingly, the feature representations at each level of grouping correlate with multiple levels of semantic meanings such as baseball players and their body parts.

We next demonstrate the contextual information of co-occurring objects encoded in our feature representations. We visualize the length-normalized average features of the `person' category region on Pascal VOC 2012 dataset using tSNE~\cite{maaten2008visualizing}. We represent each `person' feature with the co-occurring object categories, and observe that features in the similar semantic context are clustered. As shown in Fig.~\ref{fig:supp_tsne_context}, we observe clusters of similar co-occurring object categories, such as a person riding a horse (in cerise) or a bike (in green), \etc

\subsection{Visual Results on Semantic Segmentation}
\label{subsec:supp_vis_semantic}

We show some visual results of semantic segmentation on COCO-stuff and Potsdam in Fig.~\ref{fig:supp_vis_semantic}. Compared to SegSort~\cite{hwang2019segsort}, our results are more accurate and consistent.  Our predicted segmentations also preserve boundaries more precisely than the baseline.

\subsection{Unsupervised Semantic Segmentation}
\label{subsec:supp_unsup}
In the main paper, we perform unsupervised semantic segmentation by training from scratch on each dataset.  Here, we carry out experiments by following the settings used in SegSort~\cite{hwang2019segsort} and MaskContrast~\cite{van2021unsupervised}.  We summarize the quantitative results of unsupervised semantic segmentation using ImageNet-trained models on VOC in Table.~\ref{tab:supp_voc} according to IoU metric.   We demonstrate the efficacy of our method,  which outperforms SegSort~\cite{hwang2019segsort} and MaskContrast~\cite{van2021unsupervised} consistently among most categories by large margin.  Most strikingly, our method is able to capture categories with complex structures better than baselines, e.g., chair ($+5.8\%$), table ($+14.8\%$) and plant ($+10.6\%$).

\subsection{Inference Latency on Clustering Transformer}
\label{subsec:supp_latency}

We present the inference latency of different hierarchical clustering methods.  We test on a $640 \times 640$ image,  which is hierarchically partitioned into $25$, $16$, $9$ and $4$ segments.  We iterate KMeans and NCut for $30$ times.  As shown in Table~\ref{tab:speed}, our HSG imposes less runtime overhead than other clustering methods.  While major latency comes from the backbone CNN, HSG is still $17\%$ faster than KMeans.

% \section{Implementation Details}
% \label{sec:supp_details}

\subsection{Hierarchical Clustering Transformer}
\label{subsec:supp_hrchy_transformer}

We mostly follow ~\cite{carion2020end} to implement the transformer.  The detailed architecture of the clustering transformer is presented in Fig.~\ref{fig:supp_hrchy_transformer}.  The $(l+1)^{th}$-level transformer takes $X_l$ as inputs and forwards to the encoder.  The encoder contextually updates $X_l$ to $Y_l$ based on the pairwise correlation information of $X_l$.  Meanwhile, the decoder takes a set of query embeddings $Q_{l+1}$ as inputs and outputs the next-level cluster centroids.  $Q_{l+1}$ can be considered as the initial representations of next-level clusters.  As the clusterings should adapt with input statistics, we calculate the `mean' and `std' of $Y_l$, followed by $\mathrm{fc}$ layers, and sum them with $Q_{l+1}$ before inputting to the decoder.  The decoder contextually updates $Q_{l+1}$ to the next-level cluster centroids $X_{l+1}$, which become the inputs to the next-level transformer.  To calculate the clustering assignment, we do not use $X_{l+1}$ but $Z_{l+1}$, which shares the decoder layers with $X_{l+1}$ but transformed by a separate $\mathrm{fc}$ layer.  The soft clustering assignments are calculated as: $C_l^{l+1}=\text{softmax}(\frac{1}{\sqrt{m}}Y_l^\top Z_{l+1})$; $m$ is the feature dimension.

In particular, we replace $\mathrm{Layer Norm}$ with $\mathrm{Batch Norm}$.  We set number of heads to $4$ in the attention module, and use $2$ encoder (decoder) layers in each encoder (decoder) module.  We set $\mathrm{drop\_out}$ rate to $0.1$ during training.  For query embeddings $Q_l$ at level $l$, we randomly initiate and update them thru SGD.

In the clustering loss, the affinity matrix $A$ among base level feature $X_0$ is required to compute the modularity maximization loss.  We construct $A$ as a $k$-nearest neighbor (sparse) graph using the similarity of $X_0$, where the entry value is set to $1$.  $A$ is a binarized affinity matrix of a sparsified graph.  For MSCOCO/VOC/COCO-stuff/Potsdam, we set $k$ to $2$ within an image and its augmented views, respectively.  In such a manner, we encourage segment groupings across views.  On Cityscapes/KITTI-STEP, cropped patches from each image instance are less likely to overlap.  Without enforcing groupings across views, we search top $4$ nearest neighbors among views ($k=4$).

\subsection{Hyper-Parameters and Experimental Setup}
\label{subsec:supp_exp_settings}
We next describe the same set of hyper-parameters shared across different datasets, and summarize the different settings in Table.~\ref{tab:supp_hyper_parameters}.

For all the experiments, we set the dimension of output embeddings to $128$, temperature $T$ to $\frac{1}{16}$.  We apply step-wise decay learning rate policy, with which learning rate is decayed by $32\%$, $56\%$ and $75\%$ of total training epochs.  We obtain base-level grouping $G_0$ by iterating spherical KMeans algorithm over pixel-wise feature $V$ for $15$ steps and partition each cropped input to $4 \times 4$ segments.  During training not testing, $G_0$ is then refined by coherent regions generated from the OWT-UCM procedure.  For $G_1$ and $G_2$, we set $n_1$ and $n_2$ to $8$ and $4$.  The whole framework is optimized using SGD.  Notably, we only adopt rescaling, cropping, horizontal flipping, color jittering, gray-scale conversion, and Gaussian blurring for data augmentation.  All the other different settings are presented in Table.~\ref{tab:supp_hyper_parameters}.  For fair comparison with corresponding baselines, we apply different settings for training on MSCOCO and COCO-stuff.

Particularly, for MSCOCO, we adopt a two-stage learning strategy.  We first train the model with smaller crop size ($224 \times 224$) and larger batch size ($128$), then fine-tune with larger crop size ($448 \times 448$) and smaller batch size ($48$).  The models are trained and fine-tuned for $380$ and $8$ epochs.  We do not use spherical KMeans to generate image oversegmentation in the first stage of training.

For inference, we only use single-scale image.  For unsupervised semantic segmentation, we follow ~\cite{hwang2019segsort} to conduct nearest neighbor search to predict the semantic segmentation.  We apply spherical KMeans algorithm over $V$ to derive pixel grouping $G_0$ and base cluster feature $X_0$.  We search nearest neighbors using $X_0$ from the whole training dataset.   We set $n_0$--the number of centroids in $G_0$, to $6 \times 6$, $12 \times 24$ and $6 \times 12$ on Pascal/COCO-stuff/Potsdam, Cityscapes and KITTI-STEP dataset.  On Cityscapes and KITTI-STEP, we train the baselines with officially released code and test with our inference procedure.  Otherwise, we report the numbers according to their papers.

We follow \cite{hwang2019segsort} and adopt the UCM-OWT procedure~\cite{arbelaez2010contour} to generate coherent region segmentations from contours.  For MSCOCO and COCO-stuff, we follow \cite{zhang2020self} to detect edges by SE~\cite{dollar2014fast}.  The detector is first pre-trained on BSDS dataset~\cite{MartinFTM01} with ground-truth edge labels.  We start with threshold as $0.25$ to binarize the UCM, followed by OWT-UCM to generate the segmentations.  We gradually increase the threshold until the number of regions is smaller than $48$.  For Cityscapes/KITTI-STEP and Potsdam, we use PMI~\cite{isola2014crisp} to predict edges.  The detector only considers co-occurring statistics among paired colors, and does not require any ground-truth label.  The initial threshold is $0.05$ and $0.5$, which is increased until the number of regions is smaller than $1024$ and $128$.

\subsection{Unsupervised Semantic Segmentation with ImageNet-trained Models}
\label{subsec:supp_imagenet_exp}

We use PSPNet~\cite{zhao2016pyramid} based on ResNet50~\cite{he2016deep} as backbone CNN. The ResNet model is supervisedly pre-trained on ImageNet~\cite{deng2009imagenet} and fine-tuned on Pascal VOC 2012.  We follow mostly the same hyper-parameters, except that, we set batch size to $16$, crop size to $448$, base learning rate to $0.002$ and weight decay to $0.0005$.  The models are fine-tuned for $15$ epochs.  We set $\lambda_{E}$, $\lambda_{G}$ and $\lambda_{F}$ to $1.0$, $1.0$ and $0.1$, respectively.

\end{document}